\documentclass[conference]{IEEEtran}
\IEEEoverridecommandlockouts
\usepackage[english]{babel}
\usepackage{textcomp}
\usepackage{acronym}

\usepackage{cite}

\usepackage{graphicx}
\graphicspath{{./graphics/}}
\DeclareGraphicsExtensions{.pdf,.jpeg,.png}

\usepackage[caption=false]{subfig}

\usepackage{enumitem}
\usepackage{algorithmic}

\usepackage{booktabs}
\usepackage{tabularx,colortbl}
\usepackage{multirow}

\usepackage{amsmath,amssymb,amsfonts}
\usepackage{xfrac}

\usepackage{url}

\definecolor{gray}{cmyk}{0,0,0,.1} 

\def\BibTeX{{\rm B\kern-.05em{\sc i\kern-.025em b}\kern-.08em
    T\kern-.1667em\lower.7ex\hbox{E}\kern-.125emX}}

\acrodef{FRS}[FRS]{face recognition system}
\acrodef{DCNN}[DCNN]{deep convolutional neural network}
\acrodef{FOG}[FOG]{fraudulently obtain genuine}
\acrodef{ABC}[ABC]{automated border control}
\acrodef{GDPR}[GDPR]{General Data Protection Regulation}

\acrodef{2AFC}[2AFC]{two-alternative forced-choice}
\acrodef{1AFC}[1AFC]{one-alternative forced-choice}
\acrodef{mAFC}[\ensuremath{m}-AFC]{$m$-alternative forced-choice}

\acrodef{ACC}[ACC]{classification accuracy}
\acrodef{TPR}[TPR]{true positive rate}
\acrodef{TNR}[TNR]{true negative rate}
\acrodef{FPR}[FPR]{false positive rate}
\acrodef{FNR}[FNR]{false negative rate}
\acrodef{H}[H]{hit rate}
\acrodef{F}[F]{false alarm rate}

\acrodef{PAD}[PAD]{presentation attack detection}
\acrodef{BPCER}[BPCER]{bona fide presentation classification error rate}
\acrodef{APCER}[APCER]{attack presentation classification error rate}

\begin{document}

\title{Psychophysical Evaluation of Human Performance in Detecting Digital Face Image Manipulations}
\author{\IEEEauthorblockN{R. Nichols, C. Rathgeb,  P. Drozdowski, C. Busch}
\IEEEauthorblockA{\textit{da/sec -- Biometrics and Internet Security Research Group} \\
 Hochschule Darmstadt, Germany \\
\texttt{\{robert.nichols,christian.rathgeb,christoph.busch\}@h-da.de}}
}

\maketitle

\begin{abstract}
    In recent years, increasing deployment of face recognition technology in security-critical settings, such as border control or law enforcement, has led to considerable interest in the vulnerability of face recognition systems to attacks utilising legitimate documents, which are issued on the basis of digitally manipulated face images.
    As automated manipulation and attack detection remains a challenging task, conventional processes with human inspectors performing identity verification remain indispensable.
    These circumstances merit a closer investigation of human capabilities in detecting manipulated face images, as previous work in this field is sparse and often concentrated only on specific scenarios and biometric characteristics.
    This work introduces a web-based, remote visual discrimination experiment on the basis of principles adopted from the field of psychophysics and subsequently discusses interdisciplinary opportunities with the aim of examining human proficiency in detecting different types of digitally manipulated face images, specifically face swapping, morphing, and retouching.
    In addition to analysing appropriate performance measures, a possible metric of detectability is explored. Experimental data of 306 probands indicate that detection performance is widely distributed across the population and detection of certain types of face image manipulations is much more challenging than others.
\end{abstract}

\begin{IEEEkeywords}
Digital forensics, face manipulation, human performance, manipulation detection, presentation attacks, psychophysical evaluation, spoofing
\end{IEEEkeywords}

\section{Introduction}
\label{sec:introduction}
Face recognition is one of the most developed biometric technologies with endeavors dating back as far as 1966 \cite{bledsoe:1966}. In addition to algorithms, we often rely on the human ability to compare pairs of unfamiliar faces to verify identities, often in public security settings, such as border control. However, many studies have shown this to be a difficult task \cite{burton:2010} with individual performance to be widely distributed \cite{McCaffery2018}, regardless of professional training \cite{white:2014}.

This circumstance underlines vulnerabilities regarding identity verification processes, which are further exacerbated by a rather recent type of fraud which incorporates digitally manipulated face images to \ac{FOG} documents. This has led to growing interest in detection of manipulated images that are submitted in the application procedure for a new identity document - a task human inspectors are traditionally expected to perform with ease, even in difficult visual conditions. Nonetheless, initial research suggests that human examiners frequently exhibit poor performance in detecting certain types of face image manipulations \cite{robertson:2017,robertson:2018,kramer:2019}. 

In light of the potential security implications, face image manipulations present a serious threat in these operational contexts, thus warranting further research to better understand the process of detecting manipulated face images and human ability to do so. To contribute to a better understanding of this topic, we propose an interdisciplinary approach to testing and evaluating human performance to address limitations of previous works and incorporating findings within the vast body of research in the field of psychophysics \cite{GreenSwets66,KINGDOM2016,stevens:2017}. Specifically, the main contributions of this article are as follows: 
\begin{IEEEitemize}
    \item A comprehensive overview and literature survey of works that have evaluated human performance in detecting face image manipulations.
    \item Proposal of a methodological approach incorporating two well-established test procedures in the area of sensory science for this visual detection task, which better addresses limitations such as response bias and allow for a more robust evaluation of human performance. Furthermore, a computational measure to estimate task difficulty is used to facilitate interdisciplinary compatibility.
    \item An evaluation of human performance based on an expanded perceptual experiment incorporating multiple types and methods of face image manipulations to examine the capability of both novice and experienced individuals to detect these manipulations.
\end{IEEEitemize}
Our work is organized as follows. Relevant background information is outlined in Section \ref{sec:background} including face image manipulation types and processes, as well as a summary of basics and relevant methods in the area of visual psychophysics.
In Section \ref{sec:related_work}, related work regarding human detection performance of manipulated face images is comprehensively summarized. Section \ref{sec:experimental_evaluation} introduces our experimental approach and presents results thereof, before we conclude our paper with an outlook in Section \ref{sec:conclusion}.

\section{Background}
\label{sec:background}
Numerous everyday situations arise in both private and public sectors where identity and possibly corresponding characteristics (e.g. age, place of residence) must be verified to comply with local laws and ordinances, e.g. for gaining entry to a site or purchase of restricted goods. To that end, requesting a form of photo ID, confirming requirements and verifying legitimacy by comparing the submitted photo to the candidates' appearance is commonplace. Another critical example where the human ability to recognize faces is relied upon are checks on cross-border traffic as a matter of national security.
The task of comparing a passport reference image with a trusted probe image in a border control scenario is typically performed by border guards and, more and more regularly, by additional \ac{ABC} systems. 
In a routine border crossing setting, a border guard inspects the presented travel document and compares the hard copy photograph or the digital image extracted from the RFID chip to the appearance of the traveler in question, essentially comparing an unfamiliar face to a presented photograph. Studies \cite{burton:2010,dowsett:2015} have consistently shown individual performance in this task to be widely distributed across the population. Given practical and theoretical implications, determining the source of these differences has become a relevant issue, in turn prompting development of standardized tests such as the Glasgow Face Matching Test (GMFT \cite{burton:2010}) and the Cambridge Face Perception Test (CFPT \cite{Duchaine:2006}), amongst others, which have enabled the investigation of relationships within individual differences in performance across distinct recognition tasks regarding unfamiliar faces \cite{McCaffery2018}. An attack leveraging digitally manipulated face images assumes the presentation of \ac{FOG} documents to a border guard as part of the biometric verification process, highlighting the role of humans in detecting manipulations, and thus preventing illicit border crossings. As such attack attempts are assumed to be rare in relation to the number of legitimate transactions \cite{Ngan:2020}, the ability of professional inspectors and border guards to reliably detect face image manipulations remains a core interest, as does which role, if any, experience has in this context.

\subsection{Digital face image manipulations}
\label{sec:background:face_manipulation}

Traditionally, digital face editing as an image processing task has received high levels of professional and private interest, most commonly in the context of advertisement or entertainment. As such, it has recently become the focus of societal concerns, such as the spread of misinformation via social media \cite{viorela:2021}. Broadly speaking, face image manipulation involves the modification of specific facial attributes (e.g. age, sex, or hair), adding inconspicuous, yet disruptive artifacts, and swapping or morphing two distinct face images \cite{ferrara:2016, TOLOSANA:2020, 2021_Book_DigitalFaceManipulation}. Other manipulations include modification of facial expressions as well as the addition of decorative components such as glasses or tattoos and even generating fully fictitious faces \cite{Akhtar:2019}.

As automated photo and video editing tools such as deep learning-based face image manipulation software are publicly available and increasingly popular \cite{statfa:2019,statrf:2021}, the looming potential for attacks based on digitally manipulated face images has recently prompted first regulatory mitigation attempts by procedural changes\footnote[1]{\textit{Germany bans digital doppelg{\"a}nger passport photos}, Reuters (2020) \cite{reuters:2020}} in addition to increased efforts in the field of traditional media forensics \cite{stamm:2010,hany:2009}. In such a scenario, the attacker utilizes digital image manipulation techniques to either distort his own facial image to avoid identification (identity concealer) or combines two (or more) face images, aiming to successfully purport an identity other than his own (impostor). Subsequently, the attacker attempts to acquire legitimate documents with the manipulated image, which would go undetected due to the perceived similar appearance of the attacker and resulting image, ultimately enabled by the circumstance that many countries do not require the photo to be taken on-premise \cite{ferrera:2014}. An abundance of digital image manipulation techniques are available with varying levels of both required interaction and resulting quality, of which we briefly introduce three prominent examples as shown below.

\begin{figure}
    \centering
    \subfloat[]{
        \includegraphics[width=0.32\linewidth]{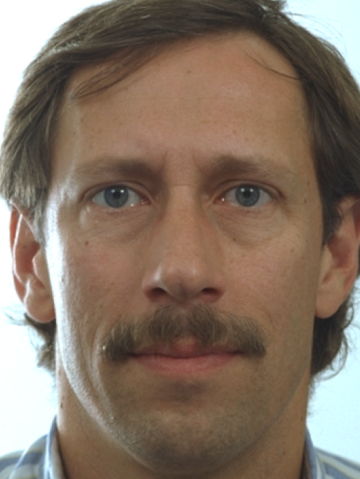}
        \includegraphics[width=0.32\linewidth]{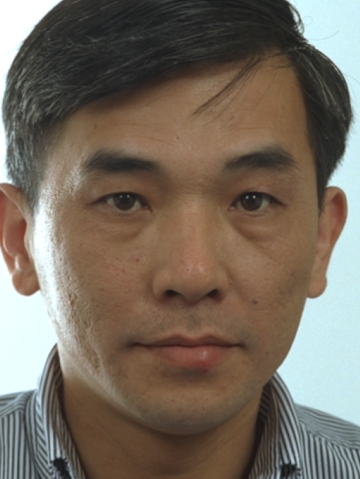}
        \includegraphics[width=0.32\linewidth]{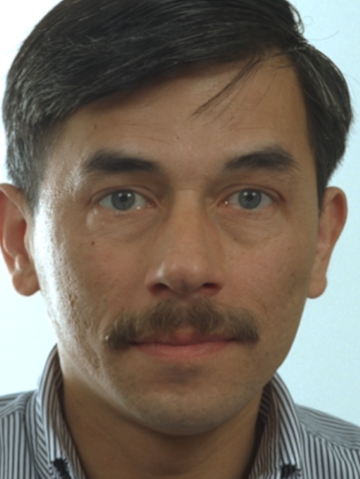}
        \label{fig:ex:fs}
    }\hfil
    \subfloat[]{
        \includegraphics[width=0.32\linewidth]{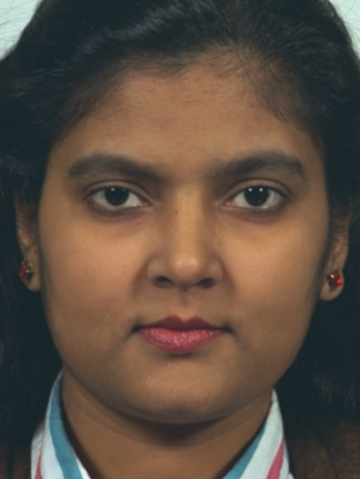}
        \includegraphics[width=0.32\linewidth]{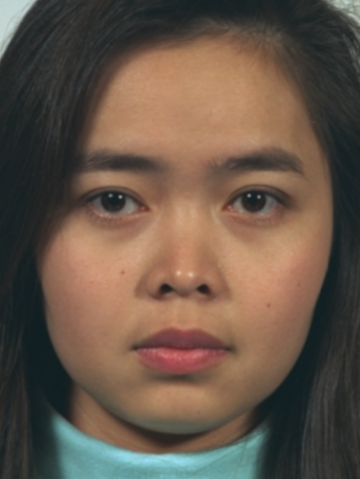}
        \includegraphics[width=0.32\linewidth]{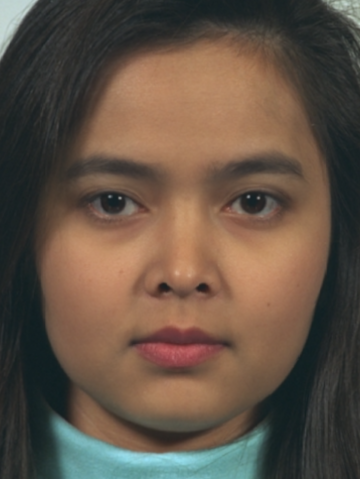}
        \label{fig:ex:m}
    }\hfil
    \subfloat[]{
        \includegraphics[width=0.32\linewidth]{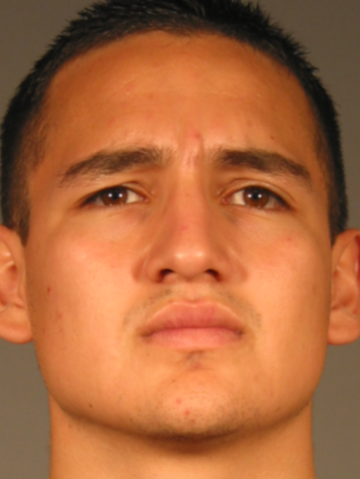}
        \includegraphics[width=0.32\linewidth]{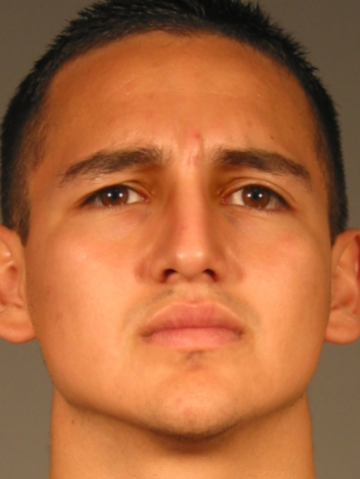}
        \label{fig:ex:rt}
    }
    \caption{Examples of bona fide source images and resulting manipulated image on the right. (a) Face swapping (b) Morphing (c) Retouching (Original image sources: Hochschule Darmstadt, and FERET [image publication permitted under fair use policy], and FRCG v.2.0 [image publication permitted under fair use policy])}
    \label{fig:ex:manipulations}
\end{figure}

\begin{description}
    \item[Face swapping] describes an image processing task with the goal of transferring the identity of a source face image onto a target face image while preserving target attributes. There are many approaches to accomplishing this task, ranging from manual image editing (e.g. Adobe Photoshop\texttrademark) to automated deep learning-based techniques, for example by changing identity information within encoded features of the target image towards identity information from source image features \cite{simswap:2020}. An example of such an automatic approach for this technique is shown in Figure \ref{fig:ex:fs}.\\

    \item[Morphing] is a multi-step process of combining two or more face images to one, which represents each contributing face to a degree relative to the respective contributions to the result \cite{ferrera:2014}. An example with two face images is shown in Figure \ref{fig:ex:m}. In principle, this is accomplished by applying generalized warping techniques and subsequent blending. In terms of required interaction, approaches range from manual \cite{wolberg:1990, beier:1992} or semi-automated \cite{Guanglai:2020} to fully automated morphing processes \cite{damer:2018}. Often an additional post-processing step (e.g. correcting contrast and brightness, manual adjustments) is carried out owing to visual artifacts in the images from previous steps.\\

    \item[Retouching] refers to a diverse set of methods of face image processing with the general goal of improving visual appearance, a task commonly performed in commercial settings such as the fashion or advertisement industry. Aside from semi-automatic methods, automatic methods which often rely on machine learning are increasingly common, with approaches ranging from skin smoothing to reduce visible imperfections \cite{lee:2009}, digital make-up application \cite{dhall:2009}, to methods for reversing age effects \cite{novskaya:2018}. See Figure \ref{fig:ex:rt} for an example of automatic attractiveness improvement by, e.g. a sharpened jawline.\vspace{1mm}
\end{description}

Recognition of face identity is critical to many everyday contexts, such as eyewitness identifications and inspection of identity documents or passports. In the past, there have been numerous reported instances of identity falsification \cite{kravets:2017,boyd:2021,costello:2018}, demonstrating a willingness of bad actors to engage in the general practice of claiming an alternative identity, also known as \emph{identity spoofing}. In the context of biometric systems, spoofing refers to a type of a presentation attack, a term describing an external attack on the biometric data capture subsystem by subverting intended operation, where the subject aims to be identified as another individual by means of presenting manipulated input to the biometric sensor \cite{ISO30107:2016}. With more and more extensive deployment of face recognition technology in security-critical areas, specifically for applications in law enforcement and border control \cite{runaway:2020,xie:2018,fastpass:2021}, face image manipulations present a serious threat in these operational contexts and to face recognition processes in general. This in turn has prompted sustained interest and work relating to detection methods for certain face image manipulations \cite{Pikoulis:2021}, in addition to aforementioned regulatory measures \cite{reuters:2020}.

\subsection{Visual psychophysics}
\label{sec:background:visual_psychophysics}
Psychophysics is concerned with relating quantifiable physical characteristics, referred to as a stimulus, to the perceptual sensation evoked in humans when they are subjected to said stimulus. Today, psychophysics mainly focuses on four perceptual primitives, namely detection, discrimination, identification, and scaling \cite{StricklandBonnieB2001TGeo}. For our work, we focus on the detection and discrimination primitives, i.e. detecting if a stimulus is present or not and the magnitude of perceptual differences. Thresholds in the psychophysical context describe a specific stimulus intensity, which is required for detection of stimulus presence \cite{KINGDOM2016}. Scaling, on the other hand, describes the differential relation of intensities of two or more given stimuli \cite{KINGDOM2016}.

\subsubsection{Psychophysical methods}
\label{sec:background:visual_psychophysics:methods}
Psychophysical methods define stimulus positioning, i.e. in which order stimuli are presented and calculation rules pertaining to the threshold or a different parameter, which can be derived from subjects' responses. Generally, these methods can be assigned to two main categories, the classic psychophysical methods, which were first developed by Fechner in 1860 \cite{fechner1860elemente}, and the more recent, adaptive methods \cite{Lawless2010, KINGDOM2016}. Below, we briefly outline three main classic methods, and broadly describe adaptive methods:

\begin{description}
    \item[Method of constant stimuli] In this method, stimuli of varied intensities are presented in random order, so that expectation errors are eliminated. The experimenter first selects a set of stimuli (e.g. five to nine) that the subject must judge in random order. The range of intensities is chosen so that the lowest intensities are never correctly recognized while the highest intensities are always recognized. After each presentation, the subject must state whether he perceived the respective stimulus or not. Thus, a psychometric function (see Section \ref{sec:background:visual_psychophysics:psychometric_function}) can be collected empirically \cite{Ehrenstein1999}. Ehrenstein \& Ehrenstein emphasize that \emph{``although the method of constant stimuli is assumed to provide the most reliable threshold estimates, its major drawback is that it is rather time-consuming and requires a patient, attentive observer because of the many trials required''} \cite{Ehrenstein1999}. A further drawback of this method is the fact that as \emph{``only stimuli near the threshold provide relevant information, many of the stimuli presented are too far away from the threshold to be of use.''} \cite{Ehrenstein1999}\\

    \item[Method of limits] In this method, the stimulus is adjusted up or down so that it is just before being perceptible or scarcely past being perceptible. The subjects must indicate the point at which they first perceive, or no longer perceive, the stimulus. In an ascending sequence, the stimulus is initially set weakly, which means the subject cannot yet perceive it, and this is increased until the subject perceives the stimulus. In a descending sequence, a high stimulus strength is set at the beginning. This is reduced until the stimulus can no longer be perceived by the subject \cite{Ehrenstein1999}. Ehrenstein and Ehrenstein found that in the \emph{``ascending and descending series of trials''} there are often \emph{``slight but systematic differences in thresholds''} \cite{Ehrenstein1999}. For this reason, the ascending and descending sequences are to be performed several times and alternately. Finally, the recorded stimulus strengths of all runs are averaged to obtain the threshold estimate \cite{Ehrenstein1999}. Zwisler proposes to include so-called catch trials, interspersed in the experiment, where no stimulus is presented. The attained frequency of perceived stimuli in such trials could be used as a corrective measure for certain response inclinations \cite{zwisler:1998}.\\
    
    \item[Method of adjustment] In this method, the stimulus is controlled by the subject and can vary continuously. Ehrenstein and Ehrenstein believe that \emph{``the simplest and quickest way to determine absolute and difference thresholds is to let a subject adjust just the stimulus intensity until it is just noticed or until it becomes just unnoticeable (in the case of measurements of the absolute threshold) or appears to be just noticeably different from, or to just match, some other standard stimulus (to measure a difference threshold).''} \cite{Ehrenstein1999}. The experimenter controls stimulus presentation and monitors the level of magnitude set by the subject. For example, a controller can be used to adjust the intensity of a sound. The subject has the task of adjusting the stimulus until it is at the threshold of perception \cite{Ehrenstein1999}. Meanwhile, \emph{``the stimulus intensity is recorded to provide an estimate of the observer's threshold.''} \cite{Ehrenstein1999}. The mean of the settings serves as an estimate for the absolute threshold \cite{Ehrenstein1999}. Some advantages of the method of adjustment are that it is more rapid than the method of limits and thus easier to use, but it may have more limited applicability because it estimates only certain points on the psychometric curve, and habituation and expectation errors are still present \cite{glasauershi:2018}.\\
    
    \item[Adaptive methods] In contrast to the previous three classic methods, for adaptive testing methods, the goal is to keep stimulus intensity close to the threshold level. They are designed so that the selection of comparison stimuli is dependent on the response behavior of the assessor. This allows for shorter procedures and thus, makes adaptive procedures particularly efficient as the authors emphasize in \cite{Ehrenstein1999}. Ehrenstein and Ehrenstein cite the staircase method as an example of such an adaptive method first used by B{\'e}k{\'e}sy (1947) \cite{bekesy1947sound}. This method is considered a modification of the threshold method. After each stimulus and the subject's response, the stimulus intensity is changed in the direction in which the stimulus threshold is assumed to be based on the response. If the subject's response is positive, the stimulus is increased; if the response is negative, the stimulus is decreased. This process is repeated until the step size reaches the preset minimum value \cite{Ehrenstein1999}. Weber believes that stimulus strengths thus \emph{``[...] approach the threshold stimulus (threshold convergence)''} and settle \emph{``around the threshold''} after a few stimulus presentations \cite{weber:1993}. 
\end{description}

\subsubsection{Response Bias}
\label{sec:background:visual_psychophysics:response_bias}
All psychophysical methods indirectly make the assumption that stimulus intensity is the sole determinant of the observer's judgment. It appears quite likely, however, that there are other factors at play which lead to a distortion of responses yet are included in that judgment. This results in a so-called response bias. Such biases can arise, for example, because the observer does not behave cooperatively in the kind of elaborate threshold determination methods encountered in settings in use, or because the certainty of the judgments can be variable. There may be observers who jump in to answer ``signal present'' at a very low certainty level and others who will only give the same verdict at a high level \cite{KINGDOM2016}. This ultimately results in different response patterns and ultimately different psychometric functions.

\subsubsection{Psychometric function}
\label{sec:background:visual_psychophysics:psychometric_function}
An important illustration of the human sensory ability with regard to detection and discrimination is the psychometric function. Such function, as described by \cite{Wichmann2001} in Equation \ref{eq:psf}, correlates a stimulus $x$ to the probabilities $P_{\Psi}$ of detecting this stimulus, where $\alpha$ and $\beta$ describe parameters for location and sensitivity regarding the base function $f$ and relate to the threshold and subject sensitivity for such a detection experiment \cite{Wichmann2001}. 
\begin{equation}
    \Psi(x;\alpha,\beta,\gamma,\lambda) = \gamma + (1-\gamma-\lambda)*f(x;\alpha,\beta)
    \label{eq:psf}
\end{equation}

\subsubsection{Test procedures}
\label{sec:background:visual_psychophysics:procedures}
Another significant aspect to consider for development of psychophysical experiments is the choice of test procedure or trial design, as used synonymously hereafter, to refer to the task itself, i.e. the number of stimuli in each trial, how they are presented and methods for data collection. The small subset of procedures described in this section are performance-based, in that they aim to measure perceptual aptitude; in other words, how well one performs at a specific visual task. In this paper, we refer to forced-choice procedures in a broad sense as all procedures that require the observer to make a choice from two or more predefined response options. This must not necessarily be dependent on the number of stimuli present in a given trial, as single stimulus trials with two response choices are equally possible and thus considered forced-choice.

\begin{description}
    \item[Yes/No] This class of procedures covers many forms of tests and is commonly employed to determine detection thresholds. Standard practice is for half of all trials to include stimuli with the target signal present and half with the target absent (i.e. noise trials). The observers task is to respond ``Yes'' or ``No'' on each trial, dependent on the target being perceived or not. The order of signal and noise trials must be random, in any case avoiding repetitious sequences of either trial type \cite{KINGDOM2016}. As a criterion-dependent procedure, Yes/No tasks are exceptionally prone to effects of response bias \cite{GreenSwets66}, as previously described. Therefore, the proportion of correct responses is often an inferior measure of observer sensitivity and the signal detection measure $d^\prime$ is generally the metric of choice for analysis \cite{KINGDOM2016}.\\

    \item[2AFC] The paired comparison, in this case specifically a \ac{2AFC} task, is considered one of the most popular procedure designs in psychophysics \cite{KINGDOM2016}. In this task, observers are presented with two stimuli A \& B, where the target signal can be either of both, and subsequently required to decide which represents the target signal. In the form we commonly refer to in this paper, both stimuli are presented side-by-side simultaneously (i.e. spatial arrangement); likewise, a form in which stimuli are displayed in temporal order is possible but is outside the scope of this work. This task is appropriate when measuring specific differences between two samples in one specific known attribute, with results providing directional information of the specified difference \cite{Lawless2010}. As a comparatively simple and intuitive task, sensory fatigue and carryover effects are reduced and the effects of response bias are considered to be minimal, given randomized balanced serving sequences. Limitations, however, can arise both when differences extend beyond the single specified sensory dimension, which could affect other sensory attributes, and when observers become frustrated due to a large number of required trials to fully sample the psychometric function \cite{Yang_2017}.\\
    
    \item[ABX] This discrimination task is a sequential match-to-sample test, where observers are, on each trial, initially presented with two reference samples A \& B, both target signal present and target absent. After the inspection phase, a third sample X is presented, which can be of type A or B and the observer is asked to indicate whether the sample X is a positive match to type A or B. As a discrimination task, the main objective is to determine if a perceptual difference exists between two sample categories \cite{Lawless2010}. The main advantages of this procedure are that no prior knowledge of the stimuli is required and that measurement of general sensory differences is enabled by allowing observers to inspect both samples, and thus independently discover attributes of difference and weight them accordingly in subsequent decisions \cite{Greenaway_2017}. Conversely, this could be a drawback as well, as the difference is not specified and observers might be influenced by overt, yet random and irrelevant attributes \cite{Lawless2010}. Other possible disadvantages could stem from the inherent reliance on memory, as opposed to strictly perceptual representations of difference and a potentially strong bias effect towards ``X = B'' in categorical perception, when reference samples are presented sequentially during the inspection phase.
\end{description}

\section{Related work}
\label{sec:related_work}
\begin{table*}[!ht]
    \renewcommand{\arraystretch}{1.1}
    \renewcommand{\thempfootnote}{\roman{mpfootnote}}
    \renewcommand{\thefootnote}{\roman{footnote}}
    \caption{Overview of relevant studies of human performance in detecting manipulated face images}
    \label{table:rw_table}
    \centering
    \footnotesize
\begin{tabular}{|lcl|cc|lc|l|}
    \hline
    \multirow{2}{*}{\textbf{Reference}}            & \multirow{2}{*}{\textbf{Manipulations}} & \multirow{2}{*}{\textbf{Stimuli}}& \textbf{Test} & \textbf{Chance}                   & \multirow{2}{*}{\textbf{Observers}} & \textbf{No.}       & \multirow{2}{*}{\textbf{Performance}}                 \\
                                                    &                                          &                   & \textbf{class\footnotemark[2]} & \textbf{probability}              &                                      & \textbf{trials}    &                                                       \\ 
    \hline
    \multirow{2}{*}{Ferrara \textit{et al.}}        &                                          & \multirow{2}{*}{10 Bona~fide}          &                                   &                                           & \multirow{2}{*}{44 (Expert)}              & \multirow{3}{*}{30}   & \multirow{2}{*}{MAR 9.57-100\%}               \\
    \multirow{2}{*}{(2016) \cite{ferrara:2016}}     & Morphing                                 & \multirow{2}{*}{20 Manipulated}        & Yes/No                            & {N/A\footnotemark[1]}                     & \multirow{2}{*}{543 (Non-expert)}         &                       & \multirow{2}{*}{FRR 0-22.27\%}                \\
                                                    &                                          &                                        &                                   &                                           &                                           &                       &                                               \\
    \hline
    \multirow{2}{*}{Makrushin \textit{et al.}}      &                                          & \multirow{2}{*}{7 Bona~fide}           &                                   & \multirow{3}{*}{N/A\footnotemark[1]}      & \multirow{3}{*}{42 (unspecified)}         & \multirow{3}{*}{30}   & MAR 44.6\%                                    \\
    \multirow{2}{*}{(2017) \cite{makrushin:2017}}   & Morphing                                 & \multirow{2}{*}{23 Manipulated}        & Yes/No                            &                                           &                                           &                       & FRR 43.64\%                                   \\
                                                    &                                          &                                        &                                   &                                           &                                           &                       & ACC 55.62\%                                   \\
    \hline
    \multirow{2}{*}{Robertson \textit{et al.}}      &                                          & \multirow{2}{*}{49 Bona~fide}          &                                   & \multirow{3}{*}{N/A\footnotemark[1]}      & \multirow{2}{*}{28 (Non-expert)}          & \multirow{3}{*}{49}   & \multirow{2}{*}{MAR$_b$ 68\%}                 \\
    \multirow{2}{*}{(2017) \cite{robertson:2017}}   & Morphing                                 & \multirow{2}{*}{245 Manipulated}       & Yes/No                            &                                           & \multirow{2}{*}{42 (Non-expert)}          &                       & \multirow{2}{*}{MAR$_g$ 21\%}                 \\
                                                    &                                          &                                        &                                   &                                           &                                           &                       &                                               \\
    \hline
    \multirow{3}{*}{Robertson \textit{et al.}}      &                                          & \multirow{3}{*}{49 Bona~fide}          & \multirow{3}{*}{Sorting}          & \multirow{3}{*}{\small\sfrac{1}{252}}     & \multirow{4}{*}{80 (Non-expert)}          & \multirow{4}{*}{32}   & HR$_b$ 48\% FA$_b$ 23-27\%                    \\
    \multirow{3}{*}{(2018) \cite{robertson:2018}}   & \multirow{2}{*}{Morphing}                & \multirow{3}{*}{245 Manipulated}       &                                   &                                           &                                           &                       & HR$_t$ 79\% FA$_t$ 7\%                        \\
                                                    &                                          &                                        &                                   &                                           &                                           &                       & HR$_g$ 71\% FA$_g$ 7\%                        \\\cline{4-5}\cline{8-8}
                                                    &                                          &                                        & Forced-choice                     & \sfrac{1}{2}                              &                                           &                       & ACC$_{tt}$ 89\%                               \\
    \hline
                                                    &                                          &                                        & \multirow{3}{*}{Sorting}          & \multirow{3}{*}{\small\sfrac{1}{252}}     & \multirow{4}{*}{80 (Non-expert)}          & \multirow{4}{*}{32}   & HR$_b$ 38-42\% FA$_b$ 30-33\%                 \\
    \multirow{3}{*}{Kramer \textit{et al.}}         &                                          & \multirow{3}{*}{127 Bona~fide}         &                                   &                                           &                                           &                       & HR$_t$ 41\% FA$_t$ 37\%                       \\
    \multirow{3}{*}{(2019) \cite{kramer:2019}}      & \multirow{2}{*}{Morphing}                & \multirow{3}{*}{245 Manipulated}       &                                   &                                           &                                           &                       & HR$_g$ 40\% FA$_g$ 29\%                       \\\cline{4-5}\cline{8-8}
                                                    &                                          &                                        & Forced-choice                     & \sfrac{1}{2}                              &                                           &                       & ACC$_{tt}$ 51.3\%                             \\\cline{4-8}
                                                    &                                          &                                        & \multirow{2}{*}{Yes/No}           & \multirow{2}{*}{N/A\footnotemark[1]}      & 40 (Non-expert)                           & 60                    & ACC 51.3-57.1\%                               \\
                                                    &                                          &                                        &                                   &                                           & 1410 (Non-expert)                         & 1                     & MAR 49\%                                      \\
    \hline
    \multirow{3}{*}{Makrushin \textit{et al.}}      &                                          & \multirow{3}{*}{11 Bona~fide}          & \multirow{4}{*}{Yes/No}           & \multirow{4}{*}{N/A\footnotemark[1]}      & \multirow{3}{*}{49 (Expert)}              & \multirow{4}{*}{30}   & FNR 15.41\%                                   \\
    \multirow{3}{*}{(2019) \cite{makrushin:2019}}   & \multirow{2}{*}{Morphing}                & \multirow{3}{*}{19 Manipulated}        &                                   &                                           & \multirow{3}{*}{230 (Non-expert)}         &                       & TNR 82.35-83.27\%                             \\
                                                    &                                          &                                        &                                   &                                           &                                           &                       & MAR 34.65\%                                   \\
                                                    &                                          &                                        &                                   &                                           &                                           &                       & TAR 77.3\%                                    \\
    \hline
    \multirow{2}{*}{Rössler \textit{et al.}}        & \multirow{2}{*}{Deepfake}                & \multirow{2}{*}{140 Bona~fide}         &                                   & \multirow{3}{*}{N/A\footnotemark[1]}      & \multirow{3}{*}{204 (Non-expert)}         & \multirow{3}{*}{60}   & \multirow{2}{*}{HR 68.69\%}                   \\
    \multirow{2}{*}{(2019) \cite{roessler:2019}}    & \multirow{2}{*}{Face swapping}           & \multirow{2}{*}{140 Manipulated}       & Yes/No                            &                                           &                                           &                       & \multirow{2}{*}{TNR 78.19\%}                  \\
                                                    &                                          &                                        &                                   &                                           &                                           &                       &                                               \\
    \hline
    \multirow{2}{*}{Wang \textit{et al.}}           &                                          & \multirow{2}{*}{100 Bona~fide}         &                                   &                                           & \multirow{3}{*}{40 (Non-expert)}          & \multirow{3}{*}{35}   & \multirow{2}{*}{ACC 53.5\%}                   \\
    \multirow{2}{*}{(2019) \cite{wang2019}}         & Retouching                           & \multirow{2}{*}{100 Manipulated}       & Forced-choice                     & \small\sfrac{1}{2}                        &                                           &                       & \multirow{2}{*}{ACC 71.1\%}                   \\
                                                    &                                          &                                        &                                   &                                           &                                           &                       &                                               \\
    \hline
    \multirow{2}{*}{Korshunov \textit{et al.}}      &                                          & \multirow{2}{*}{60 Bona~fide}          &                                   & \multirow{3}{*}{N/A\footnotemark[1]}      & \multirow{3}{*}{60 (Non-expert)}          & \multirow{3}{*}{40}   & \multirow{2}{*}{TPR 24.5-71.1\%}              \\
    \multirow{2}{*}{(2020) \cite{korschunov:2020}}  & Deepfake                                 & \multirow{2}{*}{60 Manipulated}        & Yes/No                            &                                           &                                           &                       & \multirow{2}{*}{TNR 82.2\%}                   \\
                                                    &                                          &                                        &                                   &                                           &                                           &                       &                                               \\
    \hline
    \multirow{3}{*}{Makrushin \textit{et al.}}      &                                          & \multirow{3}{*}{15 Bona~fide}          &                                   & \multirow{4}{*}{N/A\footnotemark[1]}      & \multirow{3}{*}{21 (Expert)}              & \multirow{4}{*}{40}   & TPR 65.4\%                                    \\
    \multirow{3}{*}{(2020) \cite{makrushin:2020}}   & \multirow{2}{*}{Morphing}                & \multirow{3}{*}{20 Manipulated}        & \multirow{2}{*}{Yes/No}           &                                           & \multirow{3}{*}{168 (Non-expert)}         &                       & TNR 88.47\%                                   \\
                                                    &                                          &                                        &                                   &                                           &                                           &                       & MAR 57.88\%                                   \\
                                                    &                                          &                                        &                                   &                                           &                                           &                       & FRR 23.54\%                                   \\
    \hline
    \multirow{2}{*}{Nightingale \textit{et al.}}    &                                          & \multirow{2}{*}{216 Bona~fide}         &                                   & \multirow{3}{*}{N/A\footnotemark[1]}      & \multirow{3}{*}{500 (Non-expert)}         & \multirow{3}{*}{108}  & \multirow{2}{*}{ACC 54.1\% ($d^\prime$ .21)}  \\
    \multirow{2}{*}{(2021) \cite{nightingale:2021}} & Morphing                                 & \multirow{2}{*}{108 Manipulated}       & Yes/No                            &                                           &                                           &                       & \multirow{2}{*}{ACC$_t$ 60.4\% ($d^\prime_t$ .53)}\\
                                                    &                                          &                                        &                                   &                                           &                                           &                       &                                               \\
    \hline
    \multirow{2}{*}{Zhang \textit{et al.}}          &                                          & \multirow{2}{*}{15 Bona~fide}          &                                   & \multirow{3}{*}{N/A\footnotemark[1]}      & \multirow{2}{*}{19 (Expert)}              & \multirow{3}{*}{90}   & HR 64.31-88.25\%                              \\
    \multirow{2}{*}{(2021) \cite{zhang:2021}}       & Morphing                                 & \multirow{2}{*}{75 Manipulated}        & Yes/No                            &                                           & \multirow{2}{*}{52 (Non-expert)}          &                       & TNR 79.21-97.14\%                             \\
                                                    &                                          &                                        &                                   &                                           &                                           &                       & ACC 81-86\%                                   \\ 
    \hline
    \rowcolor{gray}                                 & Morphing                                 &                                        &                                   &                                           &                                           &                       &                                            \\
    \rowcolor{gray}                                 & Face swapping                            & \multirow{-2}{*}{59 Bona~fide}         & \multirow{-2}{*}{Forced-choice}   &                                           & \multirow{-2}{*}{83 (Expert)}             &                       & \multirow{-2}{*}{ACC 75.43\% ($d^\prime$ 1.07)}\\
    \rowcolor{gray} \multirow{-3}{*}{\textbf{Ours}} & Retouching                           & \multirow{-2}{*}{64 Manipulated}       & \multirow{-2}{*}{Matching}        & \multirow{-3}{*}{\small\sfrac{1}{2}}      & \multirow{-2}{*}{144 (Non-expert)}        & \multirow{-3}{*}{50}  & \multirow{-2}{*}{ACC 62.92\% ($d^\prime$ 1.02)}\\
    \hline
\end{tabular}
\vspace{5pt}

\begin{minipage}[b]{.95\textwidth}
\footnotetext[1]{In Yes/No testing, the individually set criterion may vary, resulting in a probability not equal to 1/2 \cite{Lawless2010}}
\footnotetext[2]{see e.g. \cite{Lawless2010}}
\footnotesize{
ACC: Classification accuracy \textit{(percentage of correct answers)} \\
HR Hit Rate, TPR True Positive Rate \textit{(proportion manipulated images correctly classified as manipulated)}\\
TNR: True Negative Rate \textit{(1 - FA)}\\
FA False Alarm Rate, FPR False Positive Rate \textit{(proportion bona fide images falsely classified as manipulated)}\\
MAR: Morph Acceptance Rate, FNR: False Negative Rate \textit{(1 - TPR; 50/50 morphs, if applicable)}\\

TAR: True Acceptance Rate \textit{(proportion mated facial image pairs correctly classified as from the same source)}\\
TRR: True Rejection Rate \textit{(proportion non-mated facial image pairs correctly classified as different)}\\
FAR: False Acceptance Rate \textit{(proportion non-mated facial image pairs falsely classified as from the same source)}\\
FRR: False Rejection Rate \textit{(proportion mated facial image pairs falsely classified as different or bona fide images as manipulated)}\\

$b$ = baseline/pre-training performance\\
$t$ = post training performance\\
$tt$ = performance on training task\\
$g$ = post guidance/tips performance
}
\end{minipage}
\setcounter{footnote}{1}
\end{table*}

As the present topic is a fairly recent subject of investigation, literature is relatively sparse. Nevertheless, initial studies have indicated that humans display inadequate proficiency in detecting certain manipulated face images. However, the existing research still has problems in terms of chosen approaches, trial design and appropriate performance measures. See Table \ref{table:rw_table} for an overview of relevant studies reviewed in this section.

The face morphing attack is assumed to have the highest security impact, mainly due to the possibility of \ac{FOG} documents which would sufficiently resemble two individuals and thus subvert biometric authentication and achieve positive verification for both individuals. Therefore, human detection performance has been previously assessed almost exclusively for this type of face image manipulation.
However, given the rapid development of technology at the disposal of potential adversaries, other possible scenarios and approaches should be considered, for instance so-called deepfakes and face swapping (i.e. replacing identifying information in an image with another person's depiction) or beautification and digital make-up application. One such example is when an individual aims to avoid automatic verification altogether by means of removing identifying information from the image. This causes the process to fail and forces potentially unreliable human verification. With increasing availability, capability and consequently popularity of specialized software for face image manipulation, attacks utilizing alternate manipulation techniques become more probable.

One of few studies examining alternative manipulation methods is \cite{roessler:2019}, where the human ability to distinguish deepfake (i.e. face swapping) images from originals was assessed. The face images were sourced from YouTube videos and subsequently manipulated with both graphics-based and learning-based approaches, yielding detection performance levels below chance for more sophisticated manipulations. Another assessment of deepfakes was conducted in \cite{korschunov:2020}, where short videos rather than images were presented to participants. Additionally, the authors manually assigned stimuli to five categories, representing different levels of difficulty to detect. In line with the findings of \cite{roessler:2019}, their results also showed that participants are easily deceived by a good quality deepfake video. Human performance in \cite{wang2019} was reported at chance level for detecting face images, which were automatically manipulated primarily to improve appearance or attractiveness. Stimuli were both manually and automatically generated, with the manual editing performed by professional artists being less subtle and thus yielding higher detection rates.

Two main approaches in the discussed works are predominant for determining morph detection performance. In \cite{ferrara:2016} and for individual experiments conducted in \cite{robertson:2017,kramer:2019,makrushin:2019,makrushin:2020, nightingale:2021}, participants are tasked with comparing unfamiliar faces, given combinations of either mated, non-mated, and morphed face images, essentially determining if the two face images displayed represent the same person. Considering previous work on the unfamiliar face comparison task has consistently shown humans, including experienced individuals, to be unreliable \cite{kemp:1997,burton:2010,white:2014,dowsett:2015,wirth:2017,Weatherford:2021}; this approach possibly introduces an unintentional bias by including face recognition performance as a component in the results. Moreover, as the general relationship of performance in unfamiliar face comparison and performance in detecting manipulated face images remains unclear, the conclusiveness of such findings may be limited in this context.
The other approach as seen in \cite{makrushin:2017, robertson:2018, roessler:2019, korschunov:2020, zhang:2021} and experiments in \cite{kramer:2019, makrushin:2019, makrushin:2020, nightingale:2021} consists of a direct morph detection task comprised of bona fide and morphed face images. As this approach exclusively targets perceptual detection and discrimination, it is less affected by the aforementioned bias. Nonetheless, this approach requires very conscious decisions in terms of experiment design, as many factors can substantially influence outcome. In some cases, the chosen designs do not satisfy all requirements for the subsequent evaluation, and thus reliability of results may be limited.

In \cite{robertson:2018}, an extended version of the two-out-of-five test procedure is used, which is an appropriate method for determining if a significant, yet unspecified difference exists between two categories of stimuli \cite{BILLSON:2017}. In this extended version, participants were asked to choose manipulated images out of 10 presented face images, of which 5 were manipulated, basically corresponding to a five-out-of-ten test. The main advantage of this sorting task is the low chance probability of $\sfrac{1}{10}$ and $\sfrac{1}{252}$ for the standard and extended versions respectively. In \cite{ennis:2012}, however, the authors underline the importance of task instructions regarding this test design and reported that different terminology (e.g. \textit{same pair} or \textit{grouping}) can lead to greater proportion of incorrect responses. Taking this into account, when the previously described experiment was replicated with more sophisticated face image manipulations (i.e. harder to detect) in \cite{kramer:2019}, the task instruction was changed from originally ``Which of these images have been digitally manipulated?'' to ``Which of these images are face morphs (a blend of two faces)?'', which could be an additional factor in the reported performance reduction, despite participants' receiving specialized training. Furthermore, if the instructions or guidance include detailed descriptions, suggestions and illustrations of visual cues as e.g. \cite{kramer:2019} described, it remains unclear if the perceptual component of detecting face manipulations is discounted and a shift towards the visual search component with an additional memory task takes place. Another factor potentially facilitating such a task shift is the absence of a time limit for stimulus presentation or inspection phases in most studies, with \cite{wang2019} representing one of few exceptions with a correct implementation of a spatial \ac{2AFC} test procedure, limiting stimulus presentation time to 6 seconds on each trial. Notably, \cite{roessler:2019} utilized time limits for stimulus inspection, although as intervals were chosen at random (2, 4, or 6 seconds) for each trial to simulate real-world conditions, this choice is not common in discrimination testing, arguably due to the possibility of inter- and even intra-individual biases. 
Most studies have generally followed test procedures of the ``Yes/No'' class of psychophysical tests \cite{Lawless2010}. The specific reported procedures vary across studies, both in type and degree of conformity with the recommended protocol. Despite this design being suitable to measure detection thresholds, the limitations inherent to this task type must be considered and thus evaluated accordingly. 

For example, \cite{ferrara:2016} implements the ``Yes/No'' \ac{1AFC} ``same-different'' test procedure, in which observers are presented with combinations of the only two stimuli involved. Due to the asymmetry of the discriminands (`same' and `different') this trial design is especially prone to response bias. In this case, this is a tendency for observers to respond `Same Subject' or `Different Subject', and therefore analyzing data from such trials with signal detection measures to account for any bias would be favorable \cite{KINGDOM2016}. Furthermore, an imbalanced number of presented stimulus combinations (10 `Same Subject' and 20 `Different Subject') constitutes another source of bias.

The other common ``Yes/No'' procedure seen in present studies is the \ac{1AFC} symmetric task with one stimulus per trial. Examples for this procedure can be seen in \cite{makrushin:2019,roessler:2019,korschunov:2020, makrushin:2020, nightingale:2021, zhang:2021}, where participants were presented with a face image on each trial, belonging to either of two discriminands (``real/fake'' or ``bona fide/manipulated'', respectively). A requirement for this procedure to be considered unbiased and subsequent evaluation by \emph{<<proportion correct>>} as a performance metric as stated in \cite{KINGDOM2016}, is for both discriminands to be ``symmetric''. In terms of stimuli, this implies equivalence along all dimensions except for one, in which the difference shall be the respective opposite. This is trivial for simple visual stimuli (e.g. ``arrow left'' and ``arrow right''); however, despite defining manipulations as the opposite of genuine, face images and manipulations thereof are complex, multidimensional visual stimuli which are not easily classified as opposite in absence of objective metrics to determine the degree of divergence. An argument could be made for classifying test procedures implemented in e.g. \cite{makrushin:2017, makrushin:2019} as the ``A, not-A'' (``Yes/No Response Choice'') test; however, to comply with any of the ``A, not-A'' test variations referenced in the literature, either a training phase or sequentially presented reference sample is required, and subsequent evaluations require different statistical models \cite{Lawless2010}. Further variations of this procedure were implemented in e.g. \cite{robertson:2017} as a composite, where observers had an additional alternative as a response option, effectively representing an aggregate trial design, and in \cite{kramer:2019} (see Experiment 3) as a follow-up trial after observers performed face comparison and a decision reflection phase. Problematically for the latter layout, observers completed multiple dependent trials on the same stimuli, leading to results that are likely biased. In the instance of the former design, possible qualified models or corrective metrics of performance are undetermined.

Additionally, as for all ``Yes/No'' procedures, for minimal bias the discriminands should be presented with equal probability on each trial and as highlighted in \cite{KINGDOM2016}, it is important for the observers to be aware of this. As proper implementation of this procedure is admittedly difficult and thus, rare, treatment of this procedure as free from bias and evaluation by \emph{<<proportion correct>>} as a performance measure is inadequate and analysis of signal detection measures are preferable. Indeed, in the second experiment conducted in \cite{kramer:2019}, the aforementioned procedure design was employed with equal occurrences of discriminands (i.e. 30 manipulated, 30 bona fide stimuli in 60 trials) and subsequent analysis of signal detection measures. Nevertheless, in light of questions regarding symmetry and inherent unknown intensities (i.e. detection difficulty) of manipulated stimuli, presenting each participant with a random sample of each 30 from 60 potential stimuli, rather than each participant assessing the same set of images, could exacerbate this issue. Furthermore, no indication was made whether participants were in fact aware of discriminands appearing with equal probability.

A relatively rare choice for studies discussed in this section is the ``Forced choice'' test class, which can be described as a directed comparison of one or more pairs of stimuli, thus often implemented as (spatial-) \ac{2AFC} or \ac{mAFC} procedures, respectively. This test method is appropriate for studies in which the two (or more) sample categories are distinguished by only one specific sensory attribute, which is known a priori. Given these prerequisites, this procedure is described as efficient and powerful \cite{Lawless2010}, which speaks for this test design as an excellent choice for evaluating perceptual performance when specifying manipulation as the distinctive sensory attribute. As previously mentioned, \cite{wang2019} employed a spatial \ac{2AFC} test procedure with ``image modification'' as the discriminating sensory attribute. Furthermore, this design was applied to the morph detection training task in \cite{robertson:2018} and subsequently in \cite{kramer:2019}. As this test procedure is assumed to be less prone to response bias, typically \emph{<<proportion correct>>} is an adequate measure of performance; yet, in certain cases (i.e. positional bias), signal detection measures might be more appropriate. Noteworthy, reported training task performance of 89\% accuracy for less sophisticated morphs in \cite{robertson:2018} and 51.3\% for high quality morphs in \cite{kramer:2019}, is in line with findings in \cite{wang2019}, where 71.1\% accuracy is reported for more noticeable, manual manipulations (e.g. attractiveness) by professional artists, as opposed to 53.5\% for intentionally subtle, automatic manipulation methods.

Many of the reviewed studies used performance measures based on the \emph{proportion of correct} responses. As previously mentioned, with regard to the particular test procedure, this is a potentially unreliable measure for observer sensitivity to the target stimulus due to effects of bias or an individually set decision criterion. For the two-out-of-five test, \cite{BILLSON:2017} proposes chi-squared or binomial statistics for analysis. Since \cite{robertson:2018} and \cite{kramer:2019} employed a modified form of this procedure with five stimuli with the target signal present and five without, it is possible such recommendations do not fully extend to these procedures. In contrast, a signal detection approach as pursued, thus calculating $d^\prime$ sensitivity and criterion location $c$ for analysis of variance (ANOVA). This represents a valid course of action in the context of Yes/No tasks \cite{Stanislaw:1999, GreenSwets66}, which was appropriately followed in the remaining experiments of \cite{kramer:2019}. However, results arising from alternate test procedures and corresponding theoretical models need to be interpreted with caution, as substitute models and thus corrections \cite{Hautus:1995} and performance measures might prove more robust in terms of accuracy. Few studies use $d^\prime$ sensitivity and $c$ criterion as measures of performance for analysis, despite implementing test procedures where these would be the measures of choice. For instance, in \cite{nightingale:2021} experiments are of the Yes/No class, specifically both same-different and \ac{1AFC} symmetric test types. The authors report $d^\prime$ sensitivity and $\beta$ for response bias in addition to \emph{<<proportion correct>>}. The reported low sensitivity for detecting single morphed images is noteworthy, and likely can be ascribed to the high quality and computational selection of similar individuals for morph generation. The measure $\beta$ is a historically popular choice to represent response bias, which is based on the assumption that observers make decisions based on a likelihood ratio of obtaining a given value on either signal or noise trials. Nonetheless, arguments have been made that criterion location $c$ represents a more plausible choice to measure response bias, as it presumes a response is directly based on the decision variable \cite{Stanislaw:1999}.

The main issues and challenges in the previously discussed studies pertaining to human detection performance can briefly be summarized as follows:
\begin{IEEEitemize}
    \item \textit{Approach}: evaluating human performance in either a single or differential detection scenario, therefore weighting inclusion of face comparison task performance in results against potentially less precise results.
    \item \textit{Stimuli}: focus on one type of manipulation (morphing) in addition to unquantified differences in quality of manipulations (i.e. stimulus intensity), which greatly complicates comparison and reduces practical relevance.
    \item \textit{Test procedure}: incomplete or erroneous implementation of test procedures, in some cases modifying well studied procedures, resulting in unknown effects on robustness of findings.
    \item \textit{Performance measures}: an inadequate choice of performance measures for subsequent analysis w.r.t. the implemented test procedure may not take possible biases into account, and thus could impact reliability of findings.
\end{IEEEitemize}

\section{Experimental evaluation}
\label{sec:experimental_evaluation}
In the following section, we elaborate on the details of the experiment by giving an overview of the experiment properties, trial structure and selected stimuli, define performance measures, and briefly outline limitations in this regard. We then go on to report the results, putting our findings in perspective with regard to the context of the related works.

\subsection{Experimental setup}
\label{sec:experimental_evaluation:setup}
The experiment is designed as a web-based survey that is available at \underline{\url{https://dasec.h-da.de/unclassifyd}}. Any device with a modern web browser can be used to participate. After completing a brief registration with the possibility to self-report participant characteristics and following email confirmation to discourage improper use, access to the experiment is granted. The self-reported participant characteristics include age, gender, and relevant professional experience. The data used in this work was collected on a voluntary basis and in accordance with \ac{GDPR} \cite{eu-gdpr} provisions, thus providing required details for participants' informed consent during registration.
\subsection{Performance measures}
\label{sec:experimental_evaluation:performance_measures}
Previous studies, depending on either employing a face comparison task or detection task, mostly relied on general classification accuracies and standard performance measures of biometric systems, which often mirror common measures in sensory science. The variety of overlapping definitions make comparison across studies difficult; therefore our aim is to follow the recommended measures and models in the area of sensory science and note equivalencies with standard metrics of biometric performance where applicable. The comprehensive discussion in \cite{macmillancreelman:2004} points out various procedure designs may require different approaches for evaluation, and thus different measures or different $d^\prime$ calculations, depending on perceptual decision models.

As our work focuses on the detection of manipulations as a binary classification problem and purposefully abstains from face comparison tasks, standard metrics are employed to describe detection performance. Manipulated images are understood as positive samples (i.e. signal stimuli $s$) and bona fide images as negative samples (i.e. noise stimuli $n$). Consequently, we use the following performance measures:
\begin{description}
    \item[\Ac{ACC}] is defined as the proportion of correct responses over all trials.
    \item[\Ac{TPR} or \ac{H}] is defined as the proportion of signal stimuli (manipulated images) correctly classified as manipulated.
    \item[\Ac{FPR} or \ac{F}] is the proportion of noise stimuli (bona fide images) falsely classified as manipulated, denoted as \ac{BPCER} in the context of biometric \ac{PAD} performance \cite{ISO30107-3:2017}.
    \item[\Ac{FNR} and \ac{TNR}] are the reverse metrics of \ac{TPR} and \ac{FPR}, which can enable better comparison with previous studies. \Ac{FNR} is denoted as \ac{APCER} in the biometric \ac{PAD} performance context \cite{ISO30107-3:2017}.
\end{description}


\subsubsection{2AFC procedure}
\label{sec:experimental_evaluation:performance_measures:2afc}
Data from the \ac{2AFC} procedure is represented in a $2\times 2$ stimulus-response table as shown in Table \ref{table:2afc-st-response}. For each trial, the observer reports in which location the signal stimulus occurred, as both stimuli -- signal and noise -- are present in each trial. The two stimulus sequences, in the sense of spatial order, are denoted by angle brackets and correspond to the rows of the table. In this case, as stimuli are displayed side-by-side, $<s, n>$ represents a manipulated image in the ``A'' location to the left, and a bona fide image on the right (``B'' location). For each row, the number of hits and false alarms can be understood as proportions of the total number of corresponding trials.
\begin{table}[t]
    \renewcommand{\arraystretch}{1.2}
    \caption{Stimulus-response table relating possible stimulus sequences and possible observer responses in the 2AFC procedure}
    \label{table:2afc-st-response}
    \centering
    \small
    \begin{tabular}{cccc}
                                            &                           & \multicolumn{2}{c}{Response alternative\vspace{3mm}}                                                                                                                                    \\
\multirow{2}{*}{\rotatebox[origin=c]{90}{
    Stimulus sequence\hspace{2mm}
    }}                                      &                           & \multicolumn{1}{c}{``$s = A$''}                                 & \multicolumn{1}{c}{``$s = B$''}                                                                   \\ \cline{3-4} 
                                            & \multicolumn{1}{c|}{$<s, n>$}    & \multicolumn{1}{c|}{\begin{tabular}[c]{@{}c@{}}\\
                                                                                                \vspace{1em}
                                                                                                Hits
                                                                                            \end{tabular}}                      & \multicolumn{1}{c|}{\begin{tabular}[c]{@{}c@{}}\\
                                                                                                                                                        \vspace{1em}
                                                                                                                                                        (False alarms)\\
                                                                                                                                                    \end{tabular}}                  \\\cline{3-4} 
                                            & \multicolumn{1}{c|}{$<n, s>$}    & \multicolumn{1}{c|}{\begin{tabular}[c]{@{}c@{}}\\
                                                                                                \vspace{1em}
                                                                                                False alarms
                                                                                            \end{tabular}}                      & \multicolumn{1}{c|}{\begin{tabular}[c]{@{}c@{}}\\
                                                                                                                                                        \vspace{1em}
                                                                                                                                                        (Hits)\\
                                                                                                                                                    \end{tabular}}                  \\\cline{3-4} 
\end{tabular}
\end{table}
Observers can either respond ``$signal = A$'' (``A is manipulated'') or ``$signal = B$'' (``B is manipulated''); thus only one response column is necessary to determine remaining proportions of hits and false alarms, as the sum in each row is equal to the number of trials of the respective sequence. Hence, we arbitrarily define \textit{hit} (H) and \textit{false alarm} (F) rates as \begin{equation}
    \begin{split}
        H&=P(``s = A''|<s, n>)\\
        F&=P(``s = A''|<n, s>),
    \end{split}
    \label{eq:hf2afc}
\end{equation}
and denote this by parenthesis for the numbers in the ``$signal = B$'' response column. Generally, if observers do not favor one or the other of the alternatives a priori in the spatial 2AFC procedure, the proportion of correct responses (i.e. classification accuracy $ACC$) can be considered a performance measure relatively unaffected by bias. Although rare, this design does not preclude individual observers adopting extreme decision strategies or responding to one display position more often than another; therefore we additionally calculate $d^\prime$  
by taking into account the added difficulty compared to the Yes/No task discussed in \cite{macmillancreelman:2004}, and obtain sensitivity with \begin{equation} d^\prime = \frac{1}{\sqrt{2}}[z(H)-z(F)], \label{eq:dp2afc} \end{equation} where $z(x)$ expresses the z-score, as in the relative location where the standard normal distribution is equal to $x$.

\subsubsection{ABX procedure}
\label{sec:experimental_evaluation:performance_measures:abx}
Similarly, data arising from the ABX procedure design is represented in Table \ref{table:abx-st-response}. Observers for each trial report if they believe the stimulus (``$X$'') is a signal or noise stimulus after viewing references of both types in temporal order. In this case, the table rows correspond to both signal and noise trials with manipulated and bona fide target stimuli, respectively.
\begin{table}[t]
    \caption{Stimulus-response table relating possible target stimulus occurrence and observer responses in the ABX procedure}
    \label{table:abx-st-response}
    \centering
    \small
\begin{tabular}{cccc}
                                            &                           & \multicolumn{2}{c}{Response alternative\vspace{2mm}}                                                                                                                                    \\
\multirow{2}{*}{\rotatebox[origin=c]{90}{
    Stimulus alternative\hspace{4mm}
    }}                                      &                           & \multicolumn{1}{c}{``$S$''}                                 & \multicolumn{1}{c}{``$N$''}                                                                   \\ \cline{3-4} 
                                            & \multicolumn{1}{c|}{$X=s$}    & \multicolumn{1}{c|}{\begin{tabular}[c]{@{}c@{}}\\
                                                                                                \vspace{1em}
                                                                                                $P(S|s)$\\ Hit
                                                                                            \end{tabular}}                      & \multicolumn{1}{c|}{\begin{tabular}[c]{@{}c@{}}\\
                                                                                                                                                        \vspace{1em}
                                                                                                                                                        $P(N|s)$\\ Miss
                                                                                                                                                    \end{tabular}}                  \\\cline{3-4} 
                                            & \multicolumn{1}{c|}{$X=n$}    & \multicolumn{1}{c|}{\begin{tabular}[c]{@{}c@{}}\\
                                                                                                \vspace{1em}
                                                                                                $P(S|n)$\\ False alarm
                                                                                            \end{tabular}}                      & \multicolumn{1}{c|}{\begin{tabular}[c]{@{}c@{}}\\
                                                                                                                                                        \vspace{1em}
                                                                                                                                                        $P(N|n)$\\ Correct rejection
                                                                                                                                                    \end{tabular}}                  \\\cline{3-4} 
\end{tabular}
\end{table}
Possible responses include ``$X = signal$'' (or ``X is manipulated'') or ``$X = noise$'' (or ``X is bona fide''), thus allowing us to define \textit{hit} (H) and \textit{false alarm} (F) rates in this context as \begin{equation}
    \begin{split}
        H&=P(S|s)\\
        F&=P(S|n).
    \end{split}
    \label{eq:hfabx}
\end{equation}

Our experiment includes samples of different types of face image manipulations and each is of varying subjective quality. In particular, each manipulation category features samples which are both more and less obvious to spot. For this reason, we must assume that any given observer may continuously adopt different thresholds in view of the level of sensory evidence required for a positive decision. Therefore, we conduct a signal detection analysis as recommended in \cite{Boley:2009}, applying the \textit{differencing strategy}, which \cite{Hautus2002} indicated is likely a suitable model for ABX experiments. For our experiment, rather than looking up corresponding values for $d^\prime$ in tables as included in MacMillan and Creelman's book \cite{macmillancreelman:2004}, we calculate $d^\prime$ directly by the outlined method for differencing. First, we find the largest value of \emph{<<proportion correct>>} for constant $d^\prime$, where responding is unbiased under the assumption of equal variance, from \begin{equation}P(C)_{max}=\Phi\left[\frac{z(H)-z(F)}{2}\right],\label{eq:pcunb}\end{equation} where
$\Phi$ represents the standard normal cumulative distribution function. We can then find $d^\prime$ by inserting the previously obtained $P(C)_{max}$ into \begin{equation}P(C)_{ABX}=\Phi\left(\frac{d^\prime}{\sqrt{2}}\right)\Phi\left(\frac{d^\prime}{\sqrt{6}}\right)+\Phi\left(\frac{-d^\prime}{\sqrt{2}}\right)\Phi\left(\frac{-d^\prime}{\sqrt{6}}\right)\label{eq:pcabx}\end{equation} and solving for $d^\prime$. In line with \cite{macmillancreelman:2004}, bias was calculated as the observer criterion location $c$, which is defined as \begin{equation}c=\frac{z(H)+z(F)}{-2}.\label{eq:ccrit}\end{equation}


\subsection{Procedure}
\label{sec:experimental_evaluation:procedure}
\begin{figure}
    \centering
    \includegraphics[width=85mm]{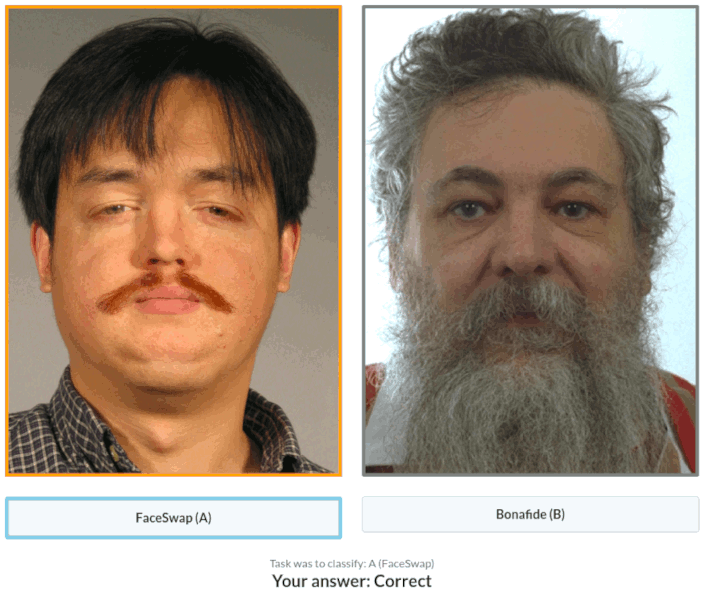}
    \caption{Example of displayed decision feedback on a 2AFC trial (Original image sources: FERET [image publication permitted under fair use policy])}
    \label{fig:trial_feedback_afc}
\end{figure}

Each trial consisted of four components: 
\begin{IEEEenumerate}
    \item a textual task description
    \item a stimulus inspection phase
    \item a decision or response step
    \item feedback display.
\end{IEEEenumerate}
Firstly, participants were made aware of the current trial procedure and objective, before inspecting the respective stimuli. Afterwards, participants were asked to indicate their assessment and to rank how confident they were in their decision on a 5-point scale from 0 (unsure) to 4 (certain).
Feedback was given immediately after each trial decision to improve task comprehension and maintain motivation, while detailed study of the stimuli during feedback was limited by a reduced display duration before progressing to the next trial. An example is given in Figure \ref{fig:trial_feedback_afc}.

\begin{figure}
    \centering
    \includegraphics[width=85mm]{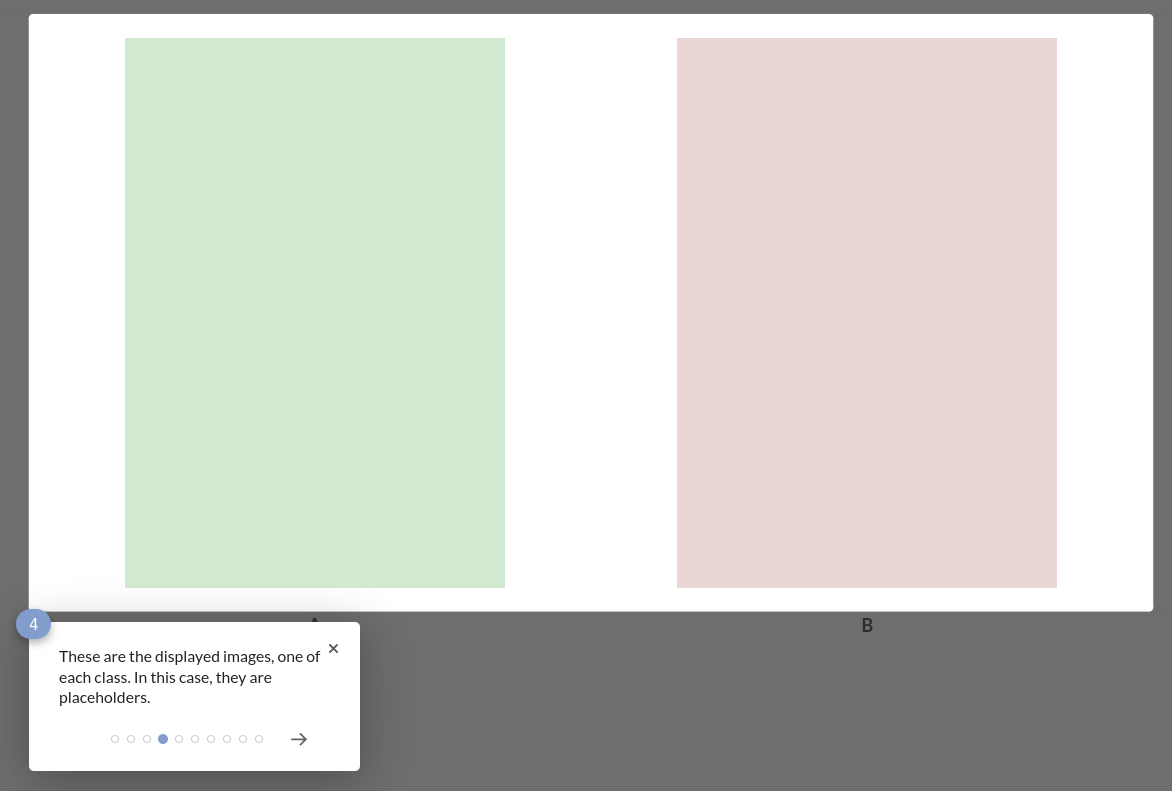}
    \caption{Interactive example trial explaining the stimulus presentation phase with placeholder stimuli to ensure participants understood and were prepared for the task at hand}
    \label{fig:example_trial_interactive}
\end{figure}

On each 2AFC trial, both stimulus alternatives were presented in a random spatial order, designated with labels \emph{A} and \emph{B}. Participants were instructed to specify the position label of the manipulated image.
Similarly, on ABX trials, two reference stimuli designated with corresponding labels \emph{Bona fide} and \emph{Manipulated} were displayed in sequential (temporal) order, followed by the target stimulus, specified as \emph{X}. Participants were instructed to specify if the last image (\emph{X}) was either manipulated or bona fide.
In visual psychophysics stimulus presentation is typically focused on very short intervals, e.g. 50 to 1000 milliseconds \cite{buracas:2005, Elze:2010}. Faces in particular represent a group of stimuli which have been found to be processed very quickly; in terms of identity representations as fast as 30 milliseconds \cite{Willis_2006, Carbon_2011}. Nevertheless, to date, detection of manipulated face images represents a novel task in a psychophysical context. We therefore chose comparatively long intervals to account for presumed greater stimulus complexity, and thus greater task difficulty. A 90-second time limit was chosen for the task description, although participants had the option to manually proceed to stimulus inspection before the timeout had elapsed. Stimuli were displayed for 8 seconds on spatial 2AFC trials and for 6 seconds each for sequential ABX trials. Participants were given 60 seconds to respond; failure to submit their assessment within the given time resulted in a nondecision and automatic progression to the next trial. Feedback display was limited to 10 seconds for previously stated reasons.
Participants each completed 23 ABX and 27 2AFC trials for a total of 50 experiment trials and two preceding instruction screens with optional interactive example trials, as illustrated in Figure \ref{fig:example_trial_interactive}. Note, the proportion of 2AFC trials was greater owing to our goal of collecting a broad sample of all stimulus intensities. As the example trials purpose was strictly to clarify tasks and methodology to minimize participant confusion, only stimulus placeholders were presented in these example trials. The experiment was based on the method of constant stimuli, a choice made in light of the advantages, despite possible drawbacks, as described in Section \ref{sec:background:visual_psychophysics}, particularly referring to accuracy and the relative procedural simplicity. As such, all trials were completed in pseudorandom order. Textual instructions and timeout parameters were uniform across all trials of the same type. Upon completion of the procedure, participants were thanked, and overall performance feedback was displayed.
\subsection{Participants}
\label{sec:experimental_evaluation:participants}
A total of 306 participants of different ages, genders, and levels of professional experience were recruited in an online campaign spanning multiple predominantly security and biometrics-focused email newsletters and social media. Based on information collected during the experimental phase, exclusion criteria was established to avoid distortion of results, as intermittent unresponsiveness on trials and delayed reactions indicated inattentive participants. Individual results were excluded when one or more of the following conditions were met:

\begin{IEEEitemize}
    \item The procedure was not completed within 6 hours
    \item Less than 23 and 27 responses were recorded for ABX and 2AFC trials, respectively
    \item The participant previously completed the experiment 
\end{IEEEitemize}

\begin{figure}
    \centering
    \includegraphics[width=85mm]{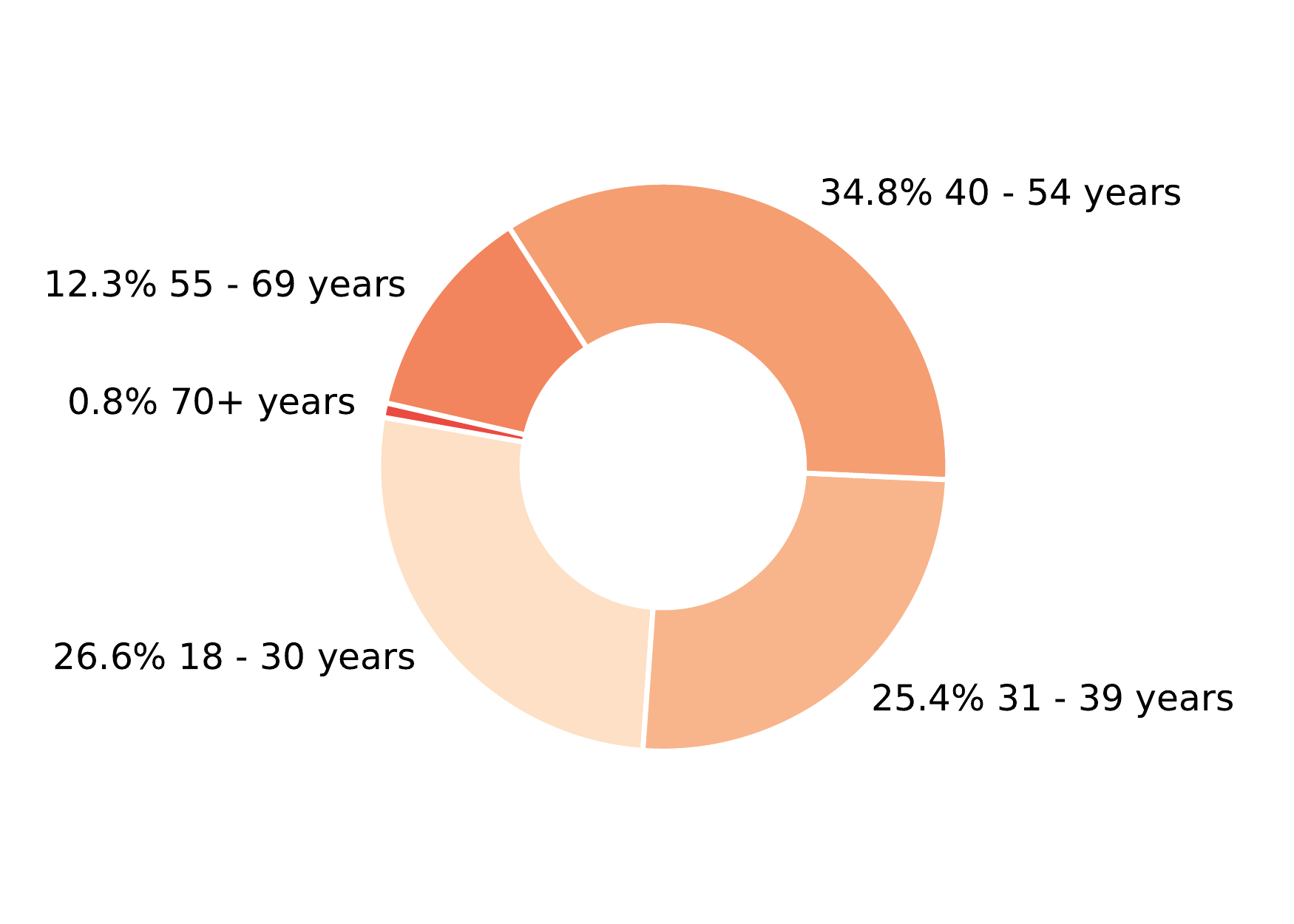}
    \caption{Participant demographics: Age distribution \hfill }
    \label{fig_age-donut}
\end{figure}

\begin{figure}[h]
    \centering
    \includegraphics[width=80mm]{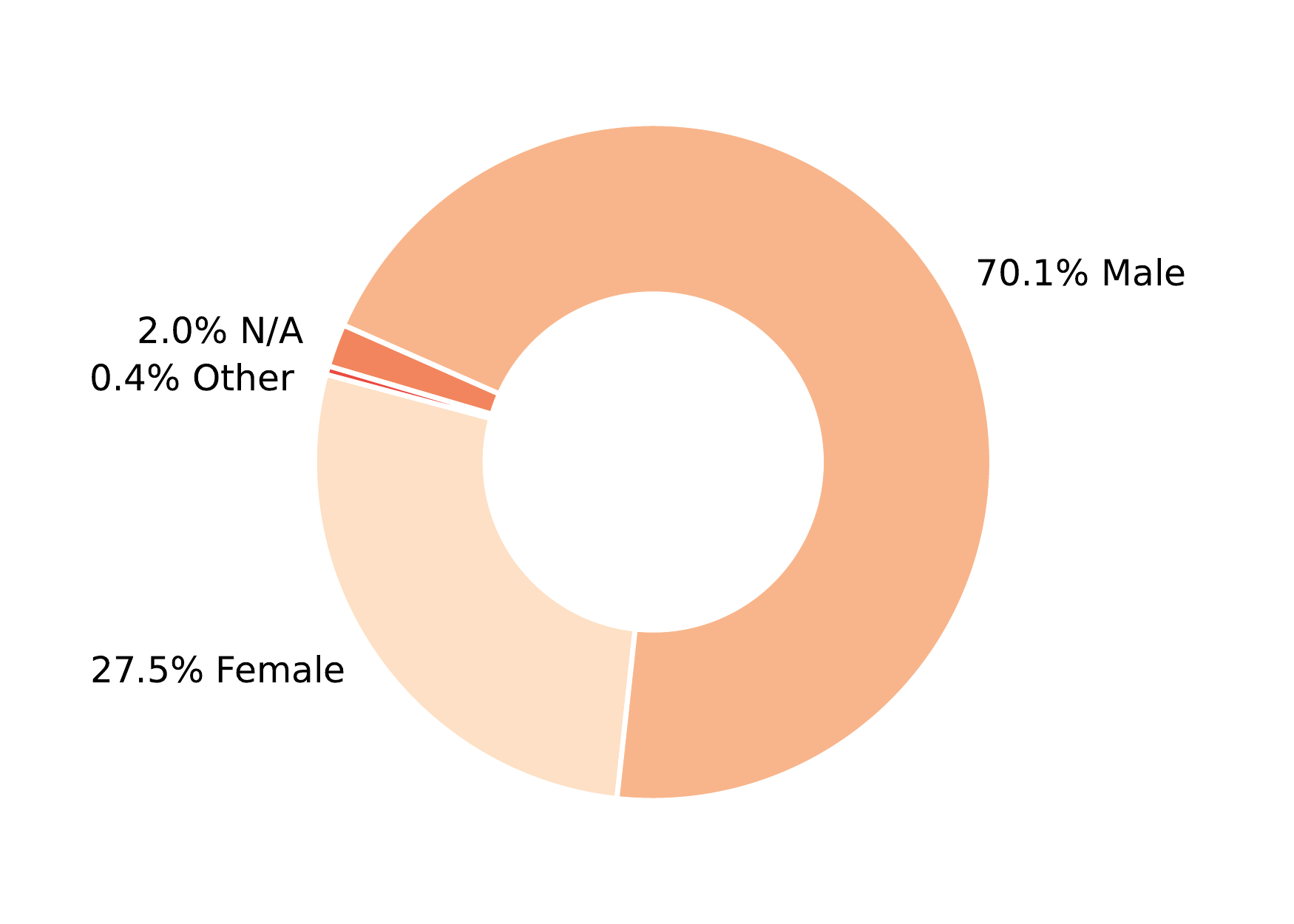}
    \caption{Participant demographics: Gender distribution \hfill }
    \label{fig_gender-donut}
\end{figure}

\begin{figure}
    \centering
    \includegraphics[width=85mm]{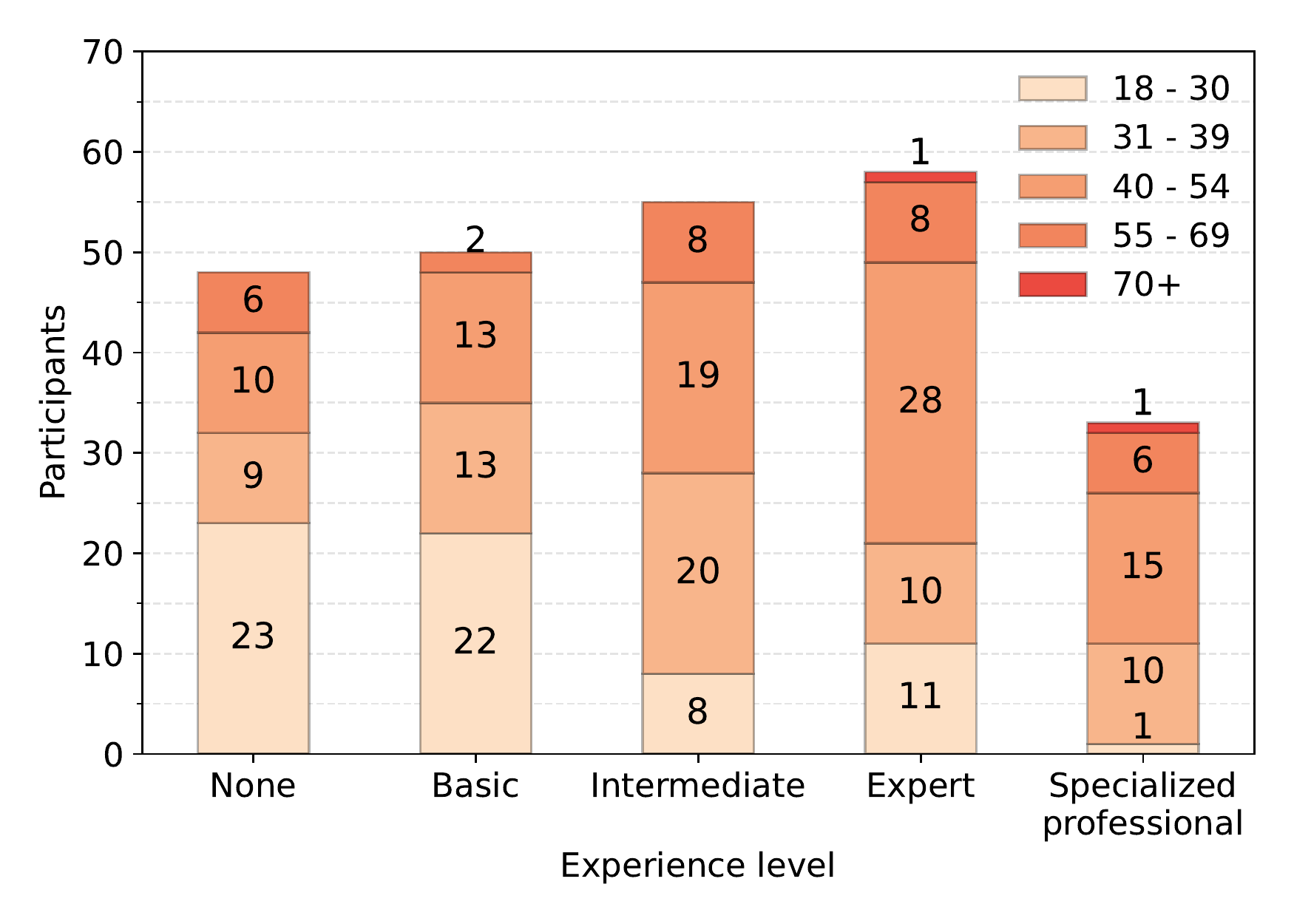}
    \caption{Participant demographics: Age distribution in relation to experience}
    \label{fig_expagedis}
\end{figure}

\begin{figure}
    \centering
    \includegraphics[width=85mm]{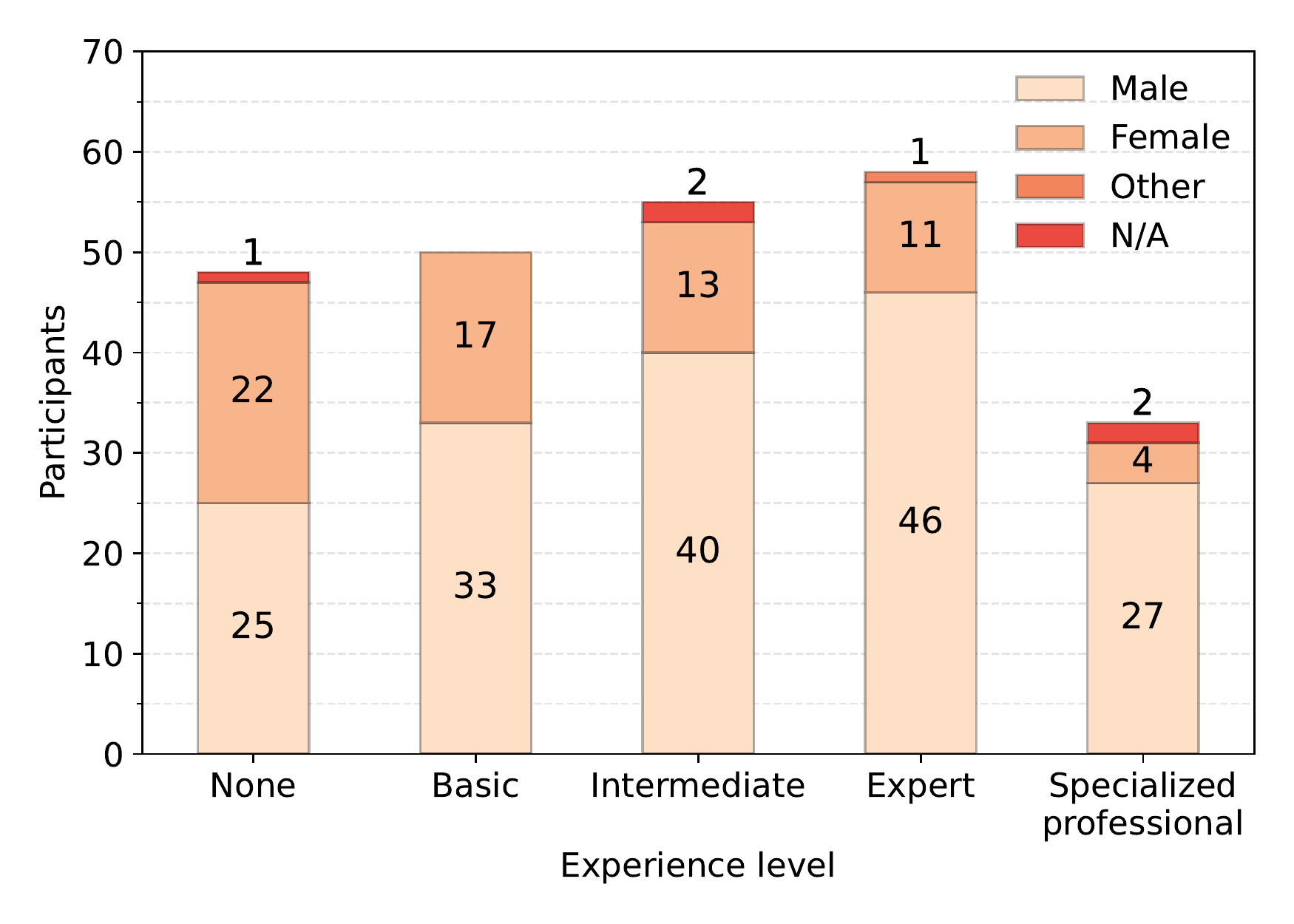}
    \caption{Participant demographics: Gender distribution in relation to experience}
    \label{fig_expgendis}
\end{figure}

Of 306 registered participants, 249 completed the full experiment. Furthermore, 22 results failed to meet remaining criteria, allowing only 227 participant results to be included in our evaluation. Two independent sample t-tests were performed to compare age and gender differences between novice (experience levels \textit{``None''} through \textit{``Intermediate''}) and expert (levels \textit{``Expert''} through \textit{``Specialized professional''}) groups. There was both a significant difference in age between novice ($N = 144$, $M = 2.13$, $SD = .998$) and expert ($N = 83$, $M = 2.71$, $SD = .957$) participants $t(225) = -4.272$, $p = .00$; as well as in gender $t(225) = 2.121$, $p = .035$. An overview of participant demographics is given in Figures \ref{fig_age-donut} and \ref{fig_gender-donut}, and characteristics related to experience in Figures \ref{fig_expagedis} and \ref{fig_expgendis}.
\begin{table}
    \renewcommand{\arraystretch}{1.2}
    \caption{Average ArcFace distances for each manipulation type and difficulty, used as an approximation of stimulus intensity or difficulty}
    \label{table:trial_target_stimuli_af}
    \centering
    \small
    \begin{tabularx}{\linewidth}{|l|Xl|c|}
        \hline
        \multicolumn{3}{|c|}{\textbf{Stimulus class}} & {\textbf{Avg. ArcFace score}} \\ \hline
        \multirow{3}{*}{Face swapping}  & \multirow{2}{*}{hard}  & Fewshotface          & 0.37 \\
                                    &                        & SimpleFS             & 0.40  \\\cline{2-4}
                                    & easy                   & SimpleFS             & 0.14 \\\hline
        \multirow{4}{*}{Morphing}      & \multirow{2}{*}{hard}  & FaceFusion           & 0.55 \\
                                    &                        & UBO                  & 0.56 \\\cline{2-4}
                                    & \multirow{2}{*}{easy}  & FaceFusion           & 0.27 \\
                                    &                        & UBO                  & 0.28 \\\hline
        \multirow{4}{*}{Retouching}    & \multirow{2}{*}{hard}  & Instabeauty          & 0.60   \\
                                    &                        & Fotorus              & 0.72  \\\cline{2-4}
                                    & \multirow{2}{*}{easy}  & Instabeauty          & 0.45  \\
                                    &                        & Fotorus              & 0.41  \\
        \hline
    \end{tabularx}
\end{table}
\subsection{Stimuli}
\label{sec:experimental_evaluation:stimuli}
\begin{figure}
    \centerline{
        \subfloat[]{\includegraphics[width=0.45\linewidth]{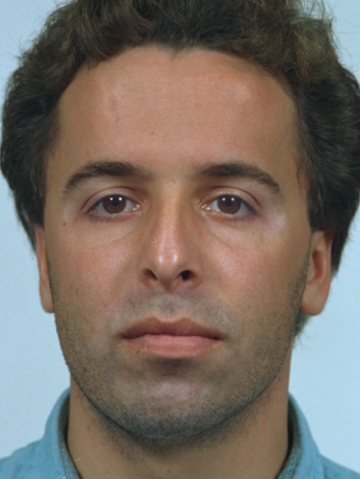}
    \label{fig:stex.ubo-hard}}
    \hfil
        \subfloat[]{\includegraphics[width=0.45\linewidth]{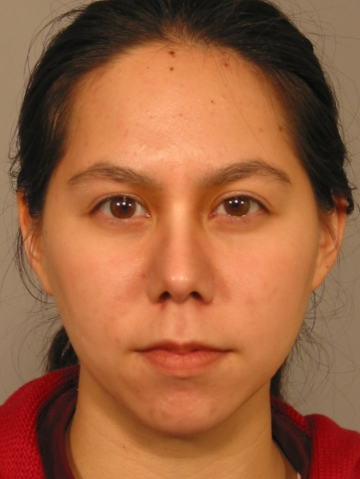}
    \label{fig:stex.fotorus-hard}}
    }
    \caption{Example stimuli of the hard category (a) Morphing: UBO (b) Retouching: Fotorus (Original image sources: Hochschule Darmstadt, and FERET [image publication permitted under fair use policy], and FRCG v.2.0 [image publication permitted under fair use policy])}
    \label{fig:stex.hard}
\end{figure}

\begin{figure}
    \centerline{
        \subfloat[]{\includegraphics[width=0.45\linewidth]{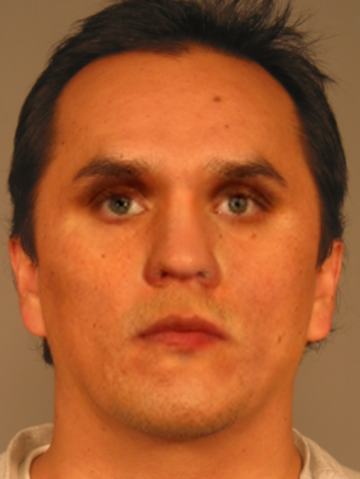}
    \label{fig:stex.fewshotface-easy}}
    \hfil
        \subfloat[]{\includegraphics[width=0.45\linewidth]{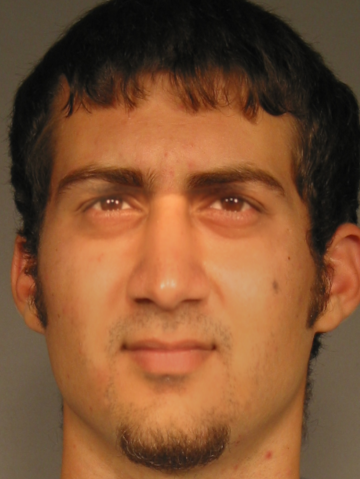}
    \label{fig:stex.facefusion-easy}}
    }
    \caption{Example stimuli of the easy category (a) Face swapping: Fewshotface (b) Morphing: FaceFusion (Original image sources: Hochschule Darmstadt, and FERET [image publication permitted under fair use policy], and FRCG v.2.0 [image publication permitted under fair use policy])}
    \label{fig:stex.easy}
\end{figure}

Stimulus facial images were derived from the FERET \cite{PHILLIPS1998295} and FRGC v.2.0 \cite{frgc} databases.
The utilized in-house database of manipulated images consisted of different automatically generated manipulations of selected images and image pairs. For each face manipulation (i.e. face swapping, morphing, retouching) two methods were used: FewShotFace\footnote{https://github.com/shaoanlu/fewshot-face-translation-GAN} and SimpleFS\footnote{https://github.com/Jacen789/simple\_faceswap}, FaceFusion\footnote{https://www.wearemoment.com/FaceFusion/} and UBO Morpher\footnote{http://biolab.csr.unibo.it/morphedfacegenerationtools.html}, FotoRus \cite{fotorus:2021} and InstaBeauty \cite{instabeauty:2021}, respectively. Thus, in total six distinct manipulation classes were available. Similarity scores were computed for each manipulated face image from a comparison with the bona fide source image, by means of ArcFace \ac{FRS}, based on a \ac{DCNN} approach utilizing Additive Angular Margin Loss \cite{deng:2019}, for which the code and pre-trained model ``LResNet100E-IR,ArcFace@ms1m-refine-v2'' were used\footnote{\url{https://github.com/deepinsight/insightface}}. Based on these scores, we further distinguished each of these classes into easy and hard categories; see Table \ref{table:trial_target_stimuli_af} for clarification.
Representative examples for both hard and easy categories are shown in Figure \ref{fig:stex.hard} and Figure \ref{fig:stex.easy}, respectively. Assuming that similarity of a bona fide source image and its manipulated counterpart can be interpreted as an indicator for detection difficulty, we surmise that low similarity scores in theory would be more noticeable, as the greater distance to the bona fide source image would suggest the image was manipulated to a higher degree. This is undoubtedly a loose approximation, as it would fail to properly reflect difficulty in terms of detection for specific manipulations and resulting signal images of high quality. As an example, considering a face swap manipulation with few visible artifacts, the similarity score may be low due to significant differences in the image. However, as the depicted face is unfamiliar to the observer, the manipulation may be hard to detect. Ten images of each class and difficulty were systematically selected for a total of 120 manipulated face images as target signal stimuli. For adequate demographic diversity within our data set, we built upon the subjective assessment of the authors. Subsequently, 60 bona fide and 60 manipulated reference face images were automatically selected to preclude any subject overlap. From this data set, image combinations balanced across manipulation types and difficulty were randomly selected as stimuli according to trial type. To summarize, our experiment included 59 bona fide and 50 manipulated target face images with an additional 14 manipulated images as reference.


\subsection{Results}
\label{sec:experimental_evaluation:results}
\begin{figure}
    \centering
    \includegraphics[width=85mm]{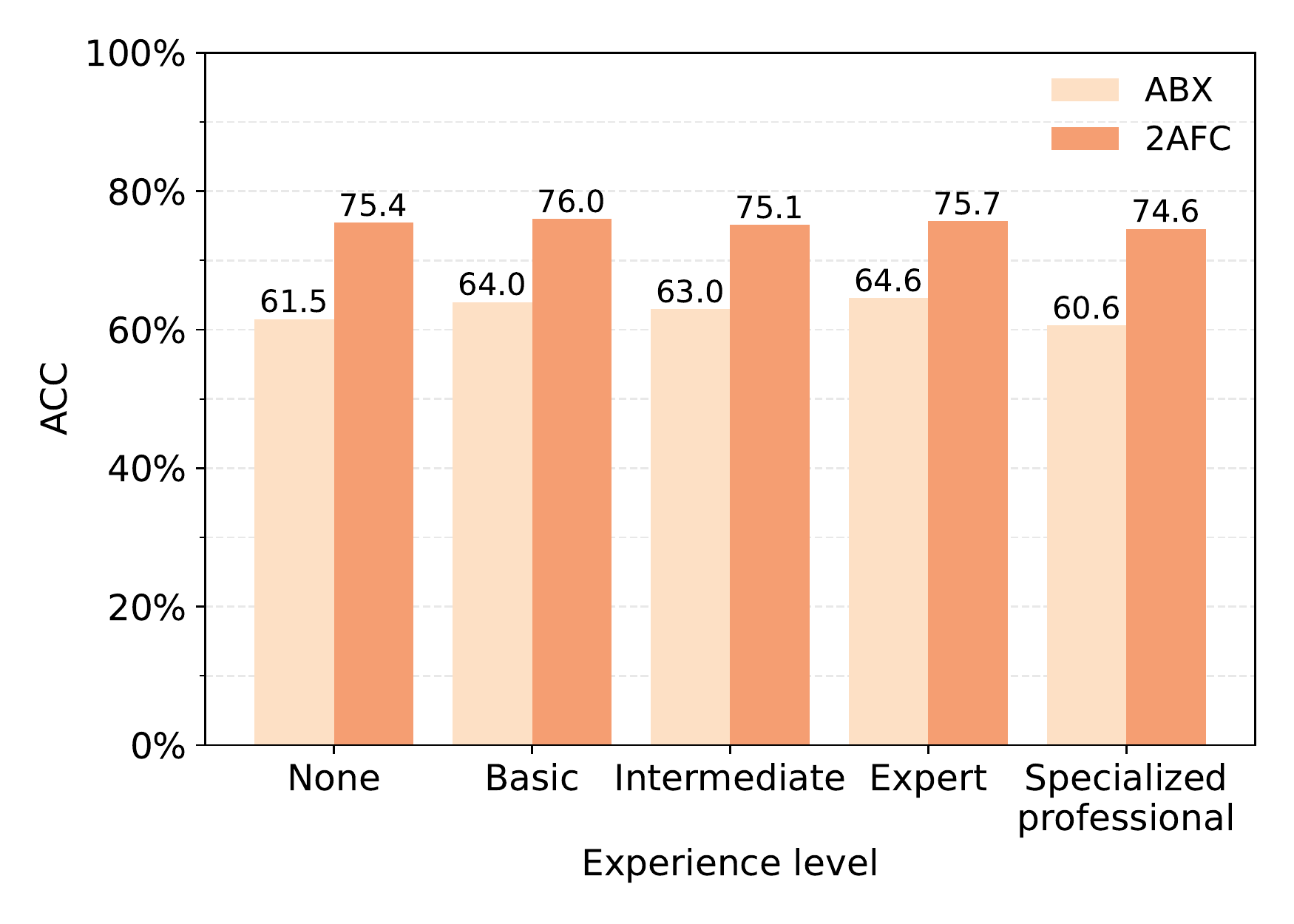}
    \caption{Average classification accuracy for each level of self-reported experience level and trial type}
    \label{fig:res:acc-experience}
\end{figure}

Results were gathered starting 2021-08-09 and the last procedure result considered was obtained on 2021-12-13. On average, participants completed the experiment in 26 minutes. 
Summary statistics for data from the experiment and corresponding measures are presented in Table \ref{table:stat_data}. 
\begin{table}
    \renewcommand{\arraystretch}{1.2}
    \caption{Summary statistics for data from all measures used in our experiment. Perceptual tasks included spatial 2AFC and sequential ABX trial procedures.}
    \label{table:stat_data}
    \centering
    \small
    \begin{tabularx}{\linewidth}{|Xrrrr|}
    \hline
    \textbf{Measure}              & \textbf{Mean}        & \textbf{Std. Dev.}& \textbf{Min.}    & \textbf{Max.}      \\ \hline
    \textbf{2AFC procedure} & \multicolumn{1}{l}{} & \multicolumn{1}{l}{} & \multicolumn{1}{l}{} & \multicolumn{1}{l|}{} \\
    \quad Accuracy (\%)           & $75.43$                & $\pm10.99$                & $40.74$                & $96.30$                  \\
    \quad H (\%)                  & $76.01$                & $\pm12.70$                 & $42.86$                & $96.43$                 \\
    \quad F (\%)                  & $25.50$                 & $\pm13.69$                & $3.85$                 & $76.92$                 \\
    \quad $d^\prime$              & $1.07$                 & $\pm0.54$                 & $-0.39$                & $2.29$                  \\ \hline
    \textbf{ABX procedure}  & \multicolumn{1}{l}{} & \multicolumn{1}{l}{} & \multicolumn{1}{l}{} & \multicolumn{1}{l|}{} \\
    \quad Accuracy (\%)           & $62.92$                & $\pm10.99$                & $34.78$                & $91.30$                 \\
    \quad H (\%)                  & 68.41                & $\pm14.63$                & 28.57                & 96.43                 \\
    \quad F (\%)                  & 45.77                & $\pm17.92$                & 11.11                & 88.89                 \\
    \quad $d^\prime$              & 1.02                 & $\pm0.95$                 & -1.50                & 3.20                    \\
    \quad $c$                     & -0.16                & $\pm2.25$                 & -13.13                & 8.13                 \\ \hline
    \end{tabularx}
\end{table}
Average classification accuracy on 2AFC trials was $75.43\%$ ($SD = 10.99\%$), and  $62.92\%$ ($SD = 10.99\%$) on ABX trials. Although participants generally performed better on the differential 2AFC task than on the sequential ABX task, mean performance was above chance level ($\sfrac{1}{2}$) in both cases. Notably, with regard to previous findings indicating individual performance in the associated unfamiliar face comparison task is widely distributed in the normal population (see e.g. \cite{bruce:2018, davis:2016, McCaffery2018}), our data suggests a similar distribution referring to the wide range of participant performance measures, though we deliberately did not replicate any previously reported face recognition tasks. This result ties well with the similar pattern of results that was obtained in the 2AFC training trials in \cite{robertson:2018}, where $SD=15\%$ and a range of $35-100\%$ was reported. Figure \ref{fig:res:acc-experience} shows the mean classification accuracy for each trial type across self-reported experience levels. In line with previous findings (e.g. \cite{ferrara:2016, makrushin:2019, makrushin:2020}), professional experience appears not to be a main factor predictive of performance in perceptually detecting digitally manipulated face images. Interestingly, for the ABX trials, the self-reported `Expert' group did perform better than examiners with no prior experience (average ACC $64.6\%$ vs. $61.5\%$). Furthermore, examiners who indicated they were specialized professionals performed worse in both ABX and 2AFC trials than those who reported only basic experience (average $ACC_{ABX} 60.6\%$ vs. $64\%$ and $ACC_{2AFC} 74.6\%$ vs. $76\%$).
\begin{figure}
    \centering
    \includegraphics[width=85mm]{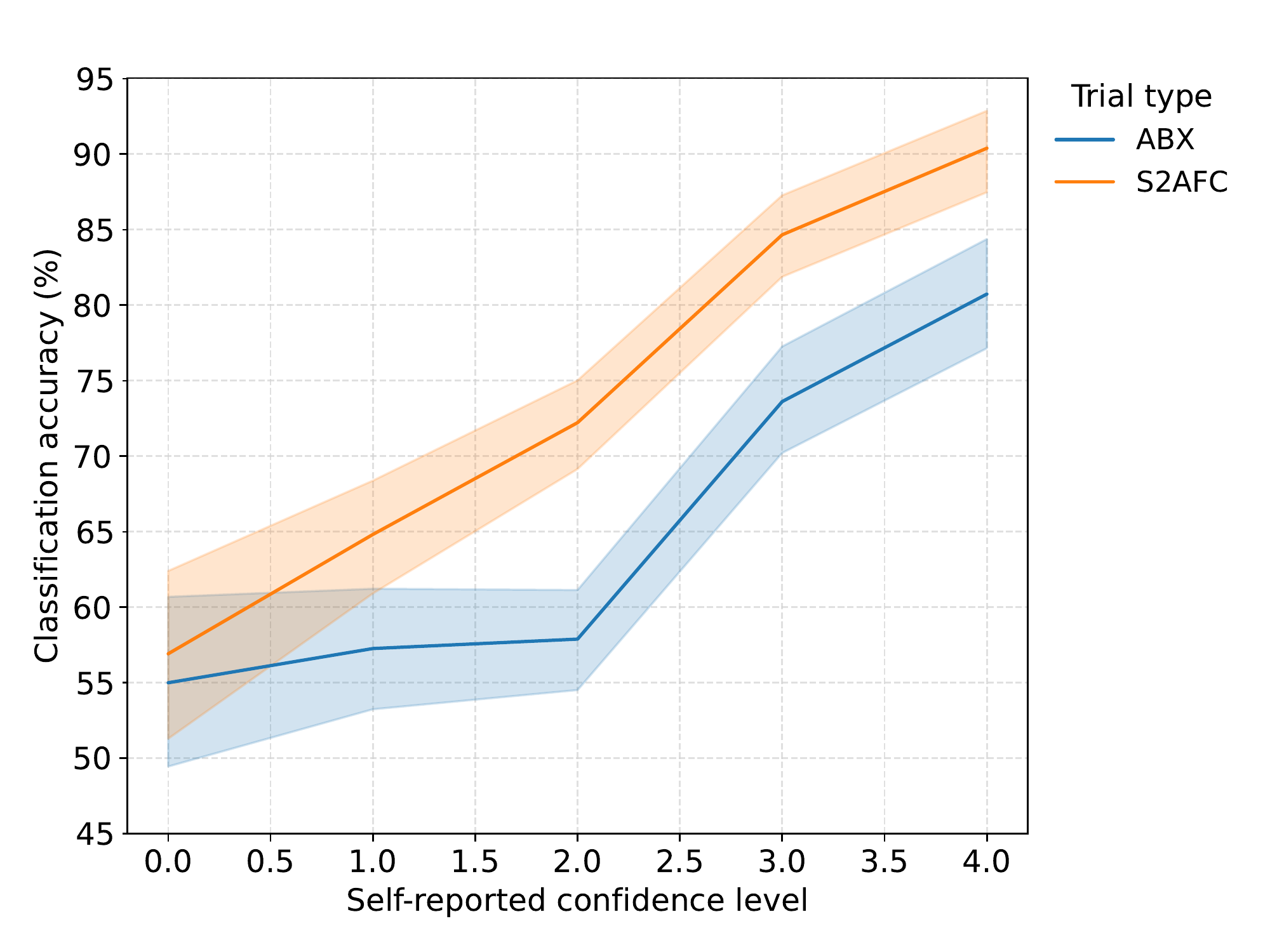}
    \caption{Average classification accuracy (with 95\% confidence interval) across self-reported confidence levels}
    \label{fig:res:acc-confidence}
\end{figure}
The correlation of decision confidence and classification accuracy is shown in Figure \ref{fig:res:acc-confidence}. Participants' confidence in having made a correct decision is reflected in accuracy. Interestingly, for the ABX trial, there appears to be a higher level of uncertainty in some cases, reflected by near chance level accuracy on the lower half of the confidence scale. This may point to participants' tendency to report confidence near the mean when they are unsure. Presumably, this is the case with ABX trials, which require a high level of attention and quick responses based on perception, as opposed to visual study of given stimuli.

\begin{figure}
    \centering
    \includegraphics[width=85mm]{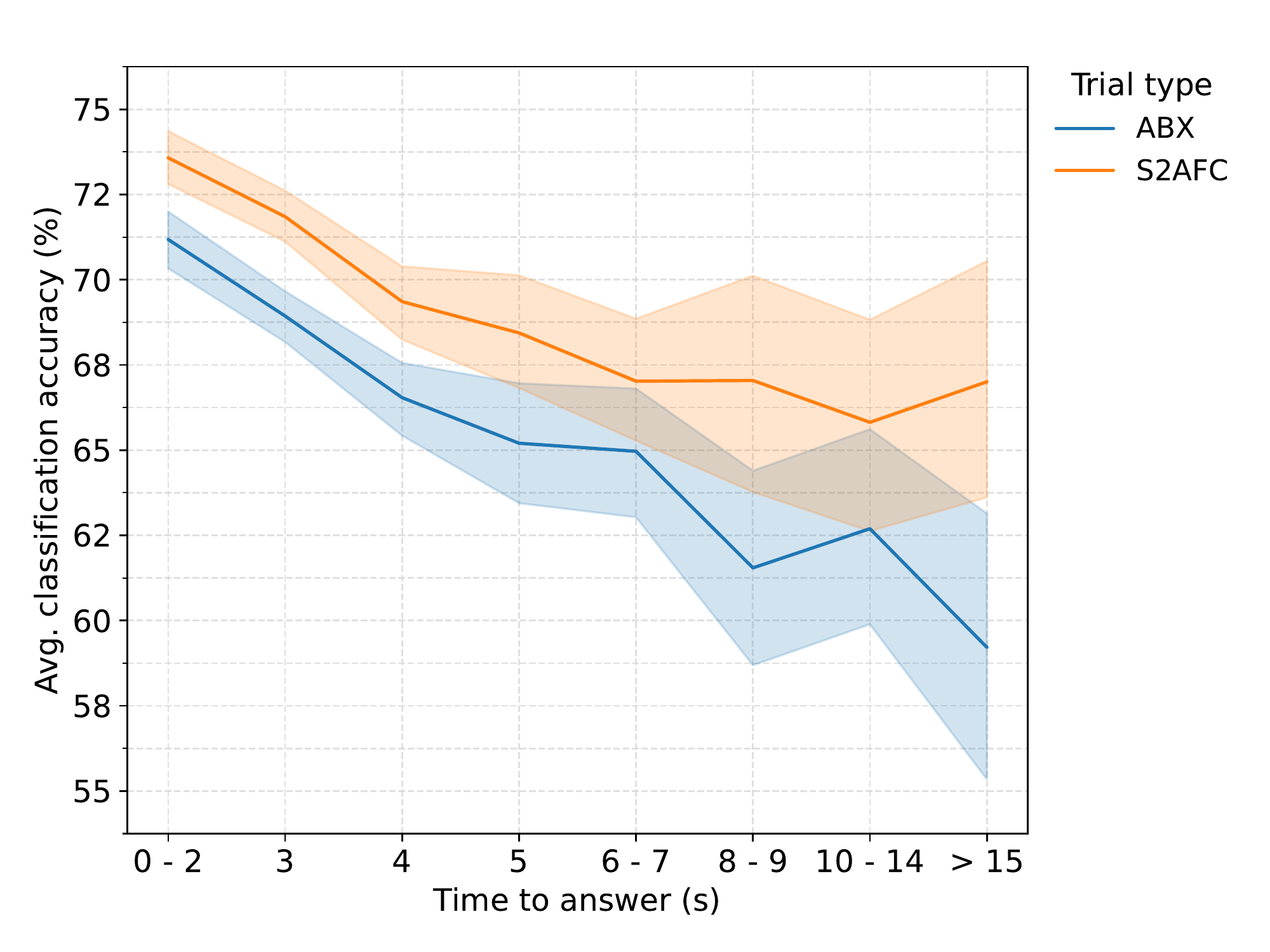}
    \caption{Average classification accuracy (with 95\% confidence interval) correlated to time in seconds participants required to submit a decision}
    \label{fig:res:acc-reaction}
\end{figure}
Conversely, classification accuracy declines the more time participants require for responding. Figure \ref{fig:res:acc-reaction} illustrates this correlation. Fast responses between 0 and 3 seconds yield the highest accuracy, indicating the participants are certain of their assessment. However, when participants require more than 8 or 9 seconds to respond, accuracy approaches levels of random guessing, in line with previous interpretation.
\begin{figure}
    \centering
    \includegraphics[width=85mm]{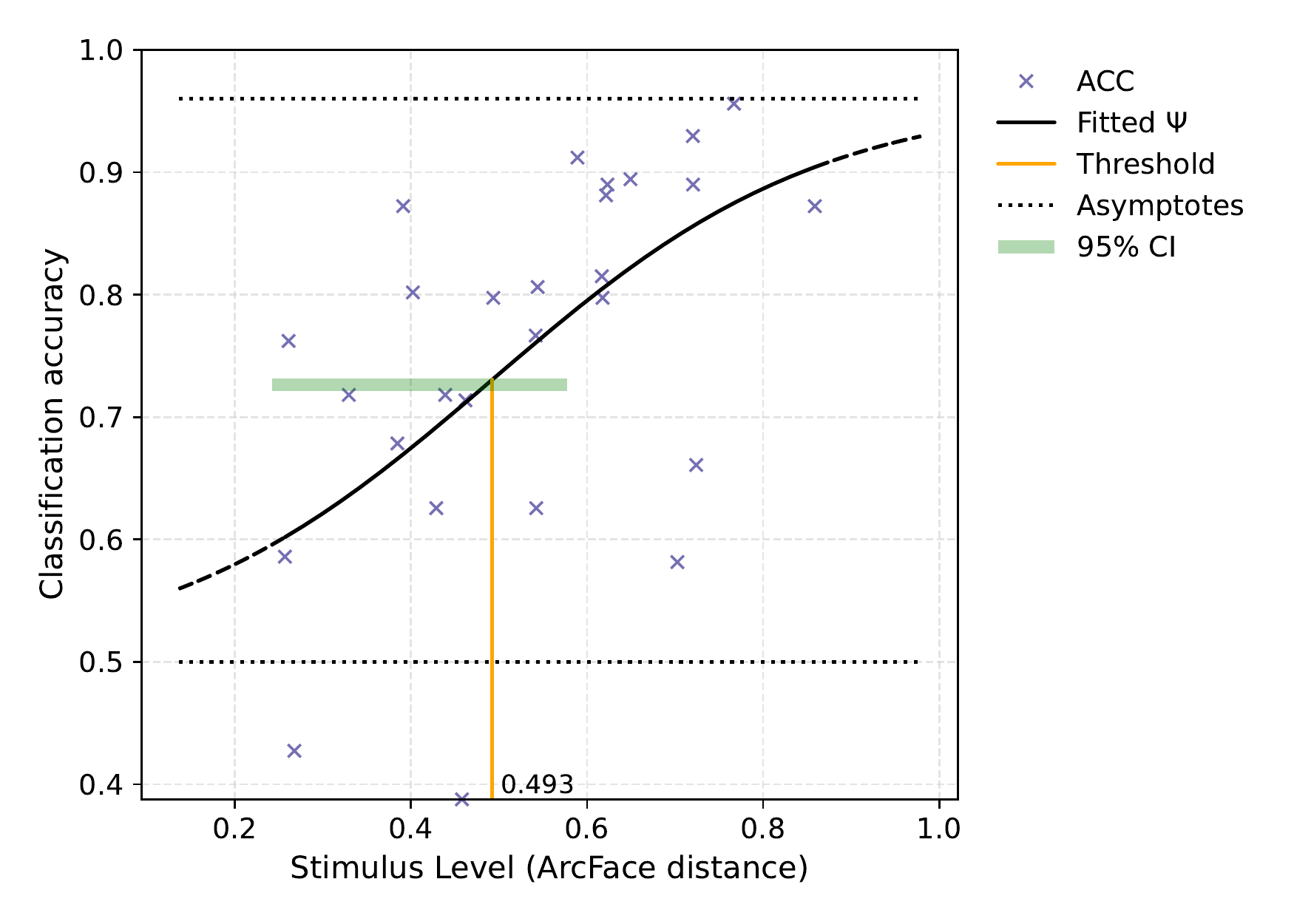}
    \caption{Estimated psychometric function based on 2AFC data. ArcFace distance scores (i.e. higher values indicate substantial differences to bona fide source images) are used as a measure of stimulus intensity.}
    \label{fig:res:2afc:psymfunc_all}
\end{figure}
\begin{figure}
    \centering
    \includegraphics[width=85mm]{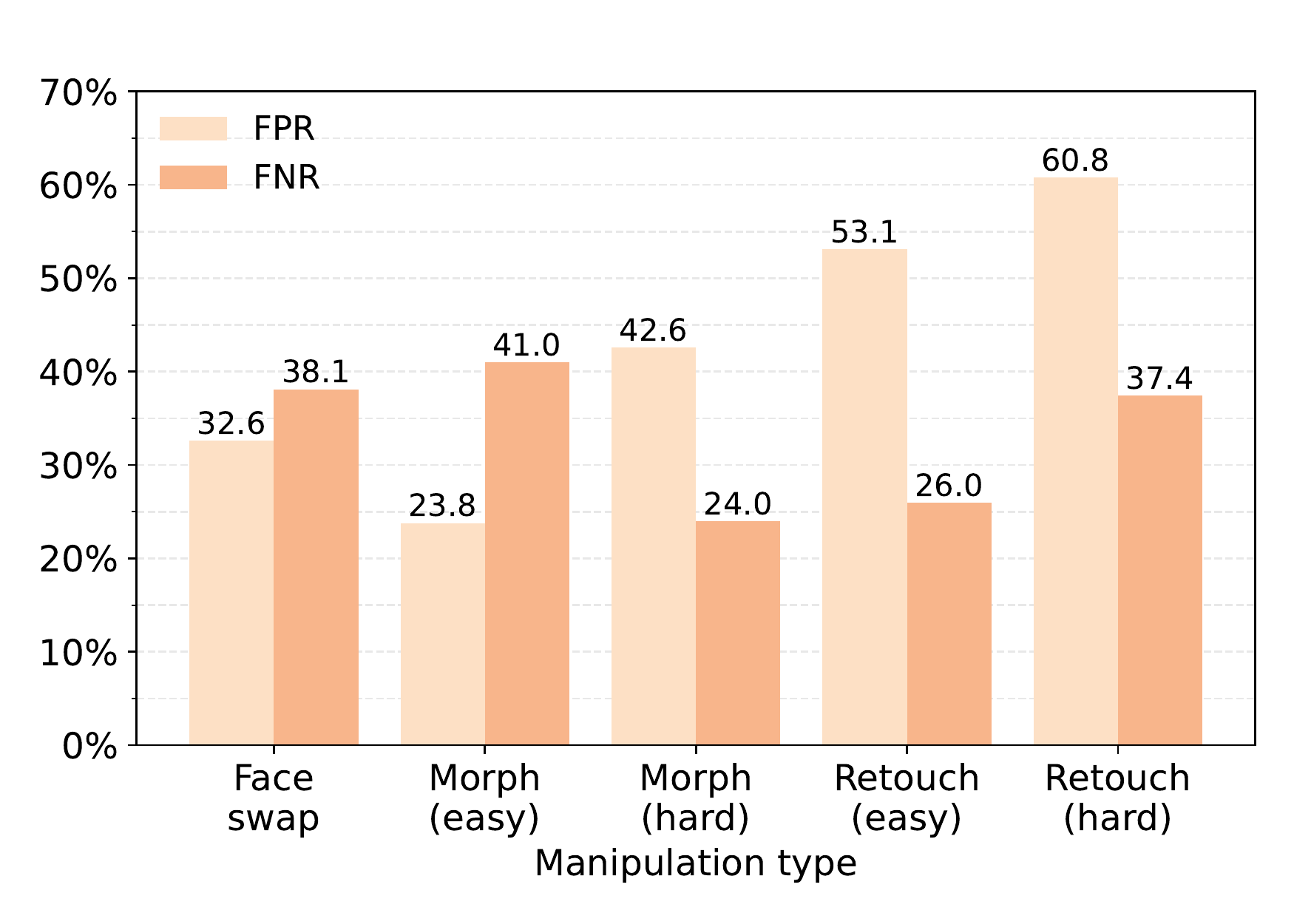}
    \caption{FPR and FNR measures for each type of manipulation in ABX trials}
    \label{fig:res:abx:signal-errors}
\end{figure}
To further explore our assignment of difficulty levels to manipulated face images as complex visual stimuli, we pursued an evaluation based on classification accuracy scores of the \ac{2AFC} trials, as these offered the widest range of hypothetical difficulty levels. The scores were computed for each trial corresponding to one manipulation type and respective difficulty by averaging the correct responses of all observers. Given these data, we conducted a Bayesian psychometric function estimation, building on the comprehensive work of \cite{schutt:2016} and utilizing the provided implementation\footnote{\url{https://github.com/wichmann-lab/python-psignifit}}. The resulting estimated psychometric function $\Psi$ is visualized in Figure \ref{fig:res:2afc:psymfunc_all}. Indeed, this result reflects our previous assumptions and findings quite well, estimating a detection threshold of $0.493$ in terms of ArcFace distance at the corresponding accuracy within the margin of error. Outlier values demonstrate the previously discussed issues regarding limited applicability of these measures of difficulty, at least for certain manipulations. Note, variance is relatively high, while the number of trials, and therefore sampled difficulties, is limited. Nonetheless, we believe that this approach is well justified to consider for further exploration and refinement.

\begin{figure*}
    \centerline{
        \subfloat[]{\includegraphics[width=85mm]{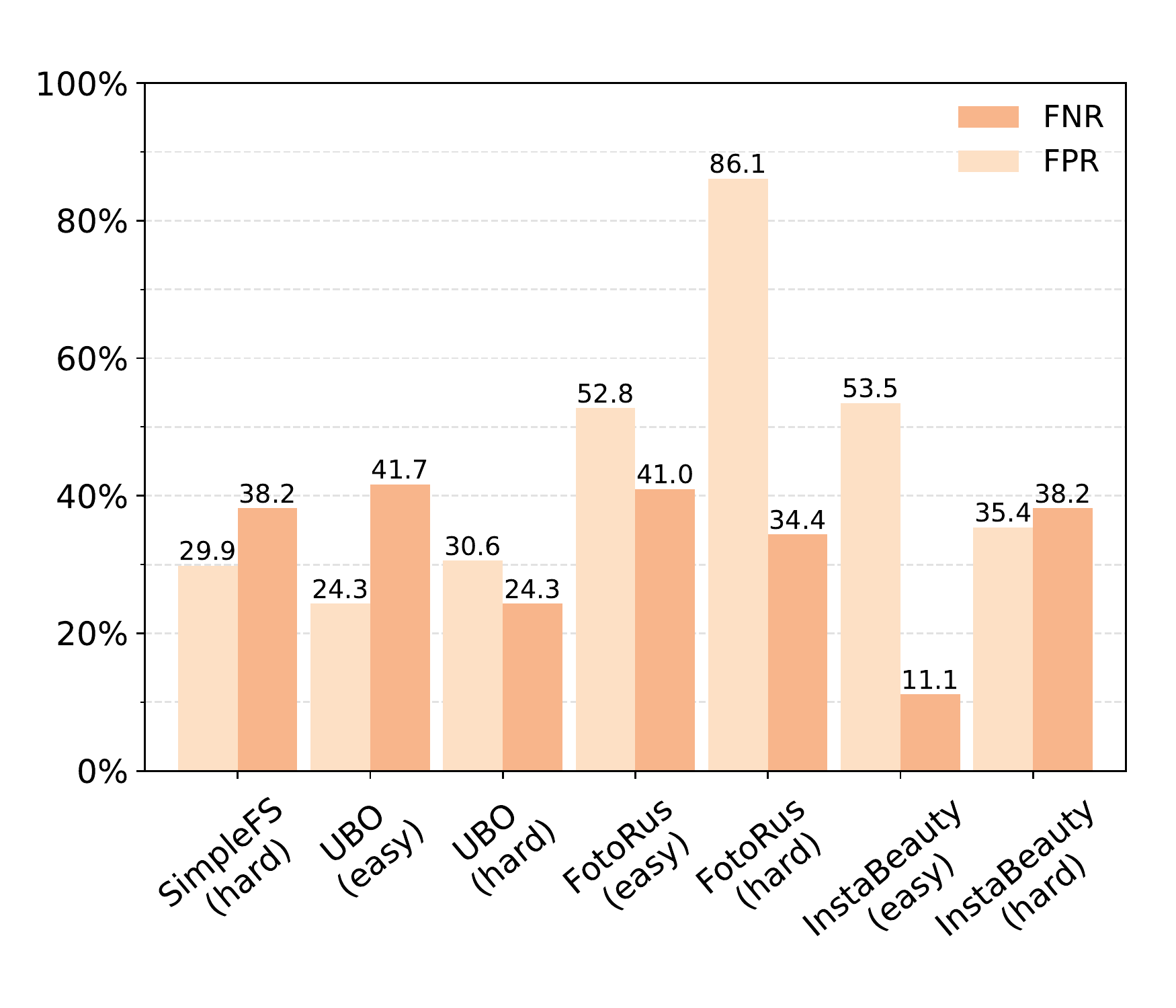}
    \label{fig:res:abx:signal_novice}}
    \hfil
        \subfloat[]{\includegraphics[width=85mm]{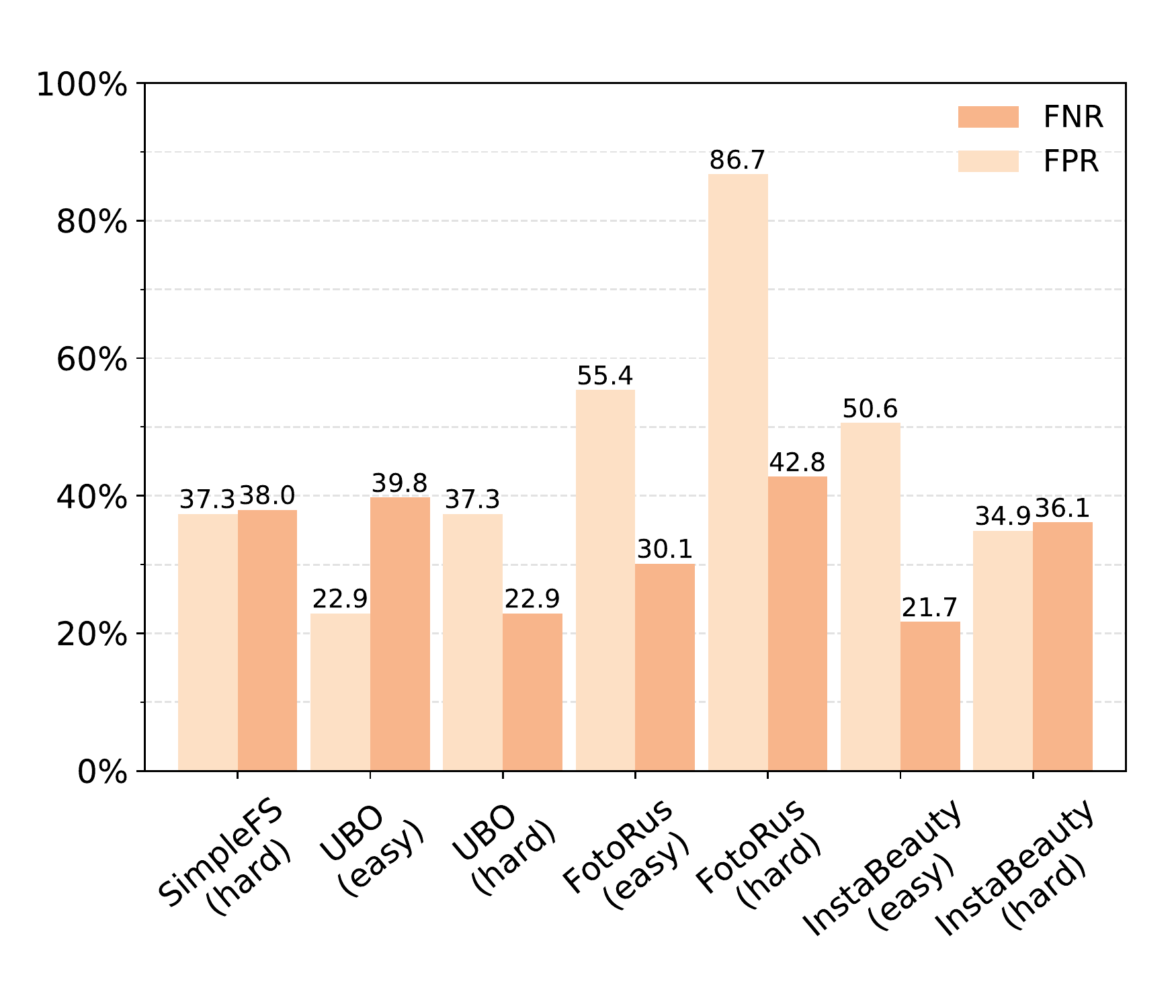}
    \label{fig:res:abx:signal_expert}}
    }
    \caption{Error rates expressed as FPR and FNR for select manipulation methods from ABX trials. Results were grouped for following experience levels:  (a) Novice (\textit{``None''} - \textit{``Intermediate''}) and (b) Expert (\textit{``Expert''} - \textit{``Specialized professional''})}
    \label{fig:res:abx:error_signal_experience}
\end{figure*}

\begin{figure}
    \centering
    \includegraphics[width=85mm]{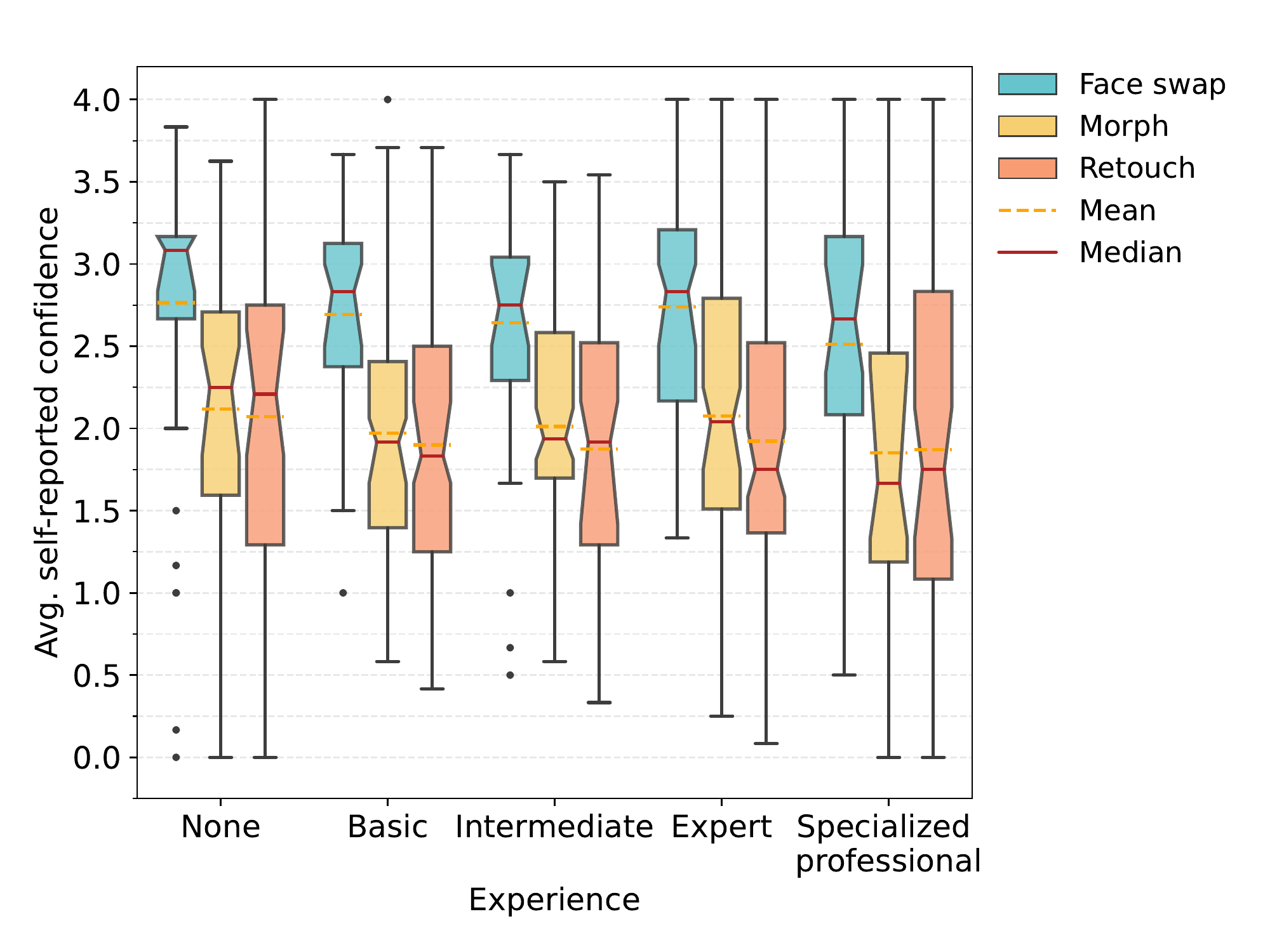}
    \caption{Confidence score as reported for each decision, grouped by experience levels for each type of manipulation, i.e. face swapping, morphing and retouching}
    \label{fig:res:info:confidence-mtypes}
\end{figure}

In terms of error rates, Figure \ref{fig:res:abx:signal-errors} shows FPR and FNR in relation to different categories of previously defined manipulation types. Generally, in line with Table \ref{table:trial_target_stimuli_af}, an increase in error rates can be observed from face swap to retouch. This confirms preliminary assumptions based on visual appearance of stimuli, but more importantly sheds light on the relationship regarding computationally determined difficulty scores. Notably, compared to the general face swap category, the presumed more difficult class of morphing experiences a reversal of FPR and FNR between the easy and hard categories. This indicates participants more willingly accept morphed images as bona fide at normalized difficulty scores, in terms of Euclidean distance, of $0.27 - 0.28$. However, when presented with morphed images with scores of $0.55 - 0.56$, the opposite appears to be the case: at this level, participants predominantly classify bona fide face images as morphs. This points to the previously mentioned limitation regarding similarity scores as detection difficulty measures of manipulated unfamiliar face images in Section \ref{sec:experimental_evaluation:stimuli}. Thus, although a relatively low similarity score might indicate substantial differences between the source and resulting manipulated image, it does not preclude a sophisticated manipulation that would not be detected by an examiner unfamiliar with the identity. Indeed, more intuitively with retouching as presumed to be the most difficult to detect manipulation, on average $53.1\%$ and $60.8\%$ of presented bona fide stimuli are incorrectly thought to be manipulated when compared to easy and hard difficulty reference stimuli, corresponding to  $0.45 - 0.41$ and $0.60 - 0.72$ difficulty scores, respectively. A more detailed breakdown of FPR and FNR error measures is presented in Figure \ref{fig:res:abx:error_signal_experience}, where select manipulation methods are shown for novice experience levels in Figure \ref{fig:res:abx:signal_novice} and for the expert experience group in Figure \ref{fig:res:abx:signal_expert}. Here, previous observations are largely reflected for both groups. Remarkably, the expert group displays a higher FPR in some cases, which supports the assumption of a lower threshold to detect manipulations, as speculated earlier. More interesting however, are the higher FNR rates, specifically for the InstaBeauty easy and FotoRus hard manipulations, indicative of a possible advantage of novice examiners. Note the high FNR rates for some manipulations, which translates to nearly a third of manipulations going undetected.
One limitation of our implementation is that our experiment scope is relatively ambitious, in that we have attempted to include many stimuli classes in addition to two trial procedures. Therefore, trials are not optimally balanced, as it was necessary to consider total experiment duration and participant fatigue, thus yielding uneven numbers of trials, which is expected to affect bias to some extent. Another limitation in terms of bias involves the issue of demographics of our sample. Due to the nature of our goal, recruiting ample numbers of good-faith participants with authentic professional experience in either biometrics or other security-related areas, we inherently reflect these demographics in terms of gender and age with regard to level of experience.
Participants were consistently more confident in their decisions regarding face swap stimuli than morph or retouch stimuli. Figure \ref{fig:res:info:confidence-mtypes} illustrates our finding across all experience levels, that participants in fact perceive detecting face swap manipulation as an easier task than e.g. morphs. Note, these data only reflect participant subjective assessment in terms of certainty of a previous decision, regardless of actual validity. Nevertheless, perceived difficulty positively correlates with stimulus difficulty measures, i.e. ArcFace distance scores, previously presented in Table \ref{table:trial_target_stimuli_af}.
\section{Conclusion}
\label{sec:conclusion}
This paper proposes a novel approach to utilizing well established methods in the area of psychophysics, implemented as a web-based experiment, to examine the human ability to perceptually detect digitally manipulated face images. We obtained an average ACC of $75.43\%$ in \ac{2AFC} trials and $62.92\%$ in the ABX trials from 227 participants. We did not find a significant difference in general performance related to the characteristic of professional experience, even given a relatively large range of options for participant self-assessment. Unexpectedly, specialized professionals were in some cases less successful than participants with little or no experience. This can be explained by two probable factors. The first is the relatively small sample size and inherent effects of gender and age bias; the second is the possibility that trained professionals have a tendency to assume manipulated images based on other internal criteria beyond basic face perception.

We have, however, shown that there are significant differences in terms of FPR and FNR among different manipulation types, suggesting different levels of difficulty can be determined. Importantly, our results indicate the use of \ac{DCNN} based \ac{FRS} to estimate detection difficulty of manipulated images might be a component in future attempts to improve the mapping of stimuli in face perception tasks. On average, performance was poor in terms of error rates, especially relatively high FPR and FNR across all experience levels on ABX trials. Nevertheless, we found individual participants who performed exceptionally well, with classification accuracy reaching $96.30\%$ and hit rates (TPR) of $96.43\%$. 

Collectively, our results are consistent with previous findings on professional experience in detecting face image manipulations and the related topic concerning face comparison and perception. This experiment adds to a growing corpus of research showing that human detection ability in this regard is limited; however, some individuals display extraordinary performance, and further study of these high-performing individuals is desirable for future work. Additionally, it will be important that future research further investigate the adequacy of different test methodologies and evaluation protocols in relation to this category of visual perception tasks.

\section*{Acknowledgment}
This research work has been funded by the German Federal Ministry of Education and Research and the Hessian Ministry of Higher Education, Research, Science and the Arts within their joint support of the National Research Center for Applied Cybersecurity ATHENE.

\bibliographystyle{IEEEtran}
\bibliography{bibliography}


\end{document}